\documentclass[a4paper,fleqn]{cas-dc}

\usepackage[numbers]{natbib}
\def\tsc#1{\csdef{#1}{\textsc{\lowercase{#1}}\xspace}}
\tsc{WGM}
\tsc{QE}
\usepackage{graphicx}
\usepackage{latexsym}
\usepackage{bm}
\usepackage{multirow}
\usepackage{colortbl}
\usepackage{bm}
\usepackage{booktabs}
\usepackage{hyperref}
\usepackage{balance}
\usepackage{subfigure}
\usepackage{enumitem}
\usepackage{bbding}
\usepackage{balance}

\begin{document}
\let\WriteBookmarks\relax
\def\floatpagepagefraction{1}
\def\textpagefraction{.001}

\shorttitle{Benchmarking Knowledge-driven Zero-shot Learning}    

\shortauthors{Yuxia Geng et al.}  

\title [mode = title]{Benchmarking Knowledge-driven Zero-shot Learning}  

%\tnotemark[<tnote number>] 

%\tnotetext[<tnote number>]{<tnote text>} 


\author[1,2]{Yuxia Geng}[orcid=0000-0002-2461-2613]

%% Footnote of the first author
%\fnmark[<footnote mark no>]

\ead{gengyx@zju.edu.cn}
\cormark[1]

%% URL of the first author
%\ead[url]{<URL>}

%% Credit authorship
%% eg: \credit{Conceptualization of this study, Methodology, Software}
%\credit{<Credit authorship details>}

\affiliation[a]{
organization={College of Computer Science \& HIC},
organization={Zhejiang University},
addressline={38 Zheda Rd}, 
city={Hangzhou},
%          citysep={}, % Uncomment if no comma needed between city and postcode
postcode={310012}, 
state={Zhejiang},
country={China}}
            
\affiliation[b]{
organization={Knowledge Engine Group, AZFT Joint Lab},
addressline={1818-2 Wenyi West Rd}, 
city={Hangzhou},
%          citysep={}, % Uncomment if no comma needed between city and postcode
postcode={311000}, 
state={Zhejiang},
country={China}}

\author[3]{Jiaoyan Chen}[]
% Email id of the second author
\ead{jiaoyan.chen@cs.ox.ac.uk}
\affiliation[c]{organization={Department of Computer Science, University of Oxford},
addressline={15 Parks Rd}, 
city={Oxford},
%          citysep={}, % Uncomment if no comma needed between city and postcode
postcode={OX1 3QD},
country={UK}}

\author[1,2]{Xiang Zhuang}[]
% Email id of the second author
\ead{zhuangxiang@zju.edu.cn}

\author[1,2]{Zhuo Chen}[]
% Email id of the second author
\ead{zhuo.chen@zju.edu.cn}

 \author[4]{Jeff Z. Pan}[]
% Email id of the second author
\ead{j.z.pan@ed.ac.uk}
\affiliation[d]{organization={School of Informatics, The University of Edinburgh},
addressline={10 Crichton St}, 
city={Edinburgh},
%          citysep={}, % Uncomment if no comma needed between city and postcode
% Informatics Forum, 10 Crichton St, Newington, Edinburgh EH8 9AB英国
postcode={EH8 9AB},
country={UK}}

\author[1,2]{Juan Li}[]
% Email id of the second author
\ead{lijuan18@zju.edu.cn}

 \author[5]{Zonggang Yuan}[]
% Email id of the second author
\ead{yuanzonggang@huawei.com}
\affiliation[e]{organization={NAIE Product Department, Huawei Technologies Co., Ltd.},
addressline={101 Software Avenue}, 
city={Nanjing},
%          citysep={}, % Uncomment if no comma needed between city and postcode
postcode={210012},
country={China}}

\author[1,2]{Huajun Chen}[]
% Email id of the second author
\ead{huajunsir@zju.edu.cn}

\cortext[1]{Corresponding author}

%% Footnote text


\begin{abstract}
External knowledge (a.k.a. side information) plays a critical role in zero-shot learning (ZSL) which aims to predict with unseen classes that have never appeared in training data.
Several kinds of external knowledge, such as text and attribute, have been widely investigated, but they alone are limited with incomplete semantics.
Some very recent studies thus  propose to use Knowledge Graph (KG) due to its high expressivity and compatibility for representing kinds of knowledge.
However, the ZSL community is still in short of standard benchmarks for studying and comparing different external knowledge settings and different KG-based ZSL methods.
In this paper, we proposed six resources covering three tasks, i.e., zero-shot image classification (ZS-IMGC), zero-shot relation extraction (ZS-RE), and zero-shot KG completion (ZS-KGC).
%For each resource, we contributed a normal ZSL benchmark and a corresponding KG containing semantics ranging from text to attributes, from relational knowledge to logical expressions.
Each resource has a normal ZSL benchmark and a KG containing semantics ranging from text to attribute, from relational knowledge to logical expressions.
We have clearly presented these resources including their construction, statistics, data formats and usage cases w.r.t. different ZSL methods.
%performance and explanations. 
More importantly, we have conducted a comprehensive benchmarking study, with two general and state-of-the-art methods, two setting-specific methods and one interpretable method.
We discussed and compared different ZSL paradigms w.r.t. different external knowledge settings, and found that our resources have great potential for developing more advanced ZSL methods and more solutions for applying KGs for augmenting machine learning.
%Extensive evaluations show us the great potential of our resources in advancing zero-shot learning. \todo{[We need a sentence to show the significance of our results.]}
% how they can be utilized with cases in evaluating ZSL methods' performance and explanations.
All the resources are available at \url{https://github.com/China-UK-ZSL/Resources\_for\_KZSL}.
\end{abstract}


% \begin{highlights}
% \item 
% \item 
% \item 
% \end{highlights}

% Keywords

\maketitle

% Main text
\section{Introduction}\label{sec:introduction}
Supervised learning has achieved great success in many domains such as natural language processing and computer vision.
Its methods often require a large number of labeled training samples to achieve good performance, following a closed world assumption.
Namely, they predict with classes that have appeared in the training stage (i.e., seen classes).
However, in many real-world applications, new classes always emerge, and it often costs too much computation, human labour and time to address these new classes by collecting labeled samples and training the model from scratch.
To this end, Zero-shot Learning (ZSL), which aims at predicting with classes that have no training samples (i.e., unseen classes), was proposed and has been widely investigated in the past decade \cite{wang2019survey,xian2018zero,chen2021knowledge}.

Since no labeled samples are given for unseen classes, existing ZSL methods usually rely on \textit{external knowledge} (a.k.a. \textit{side information}) which describes prior semantic relationships between classes.
They follow some paradigms to utilize these external knowledge to transfer data and/or models from seen classes to unseen classes.
For example, one classic paradigm is mapping-based which first embeds all the classes with their external knowledge, then (jointly) maps the class embeddings and the sample features into one common space where testing samples can be matched with classes by measuring distances with metrics such as the Cosine similarity.
%Existing ZSL methods usually learn a knowledge transfer model in the following paradigm.
%They collect and/or construct \textit{external knowledge} (a.k.a. \textit{side information}) which describes prior semantic relationships between classes, 
%embed all the classes with the external knowledge, 
%establish a mapping of inter-class relationship from the class embedding space to the sample space, and transfer model parameters (e.g., features) learned from samples of seen classes to a new model that can predict for unseen classes.
Widely investigated external knowledge includes class textual information (e.g., names and descriptions) \cite{qin2020generative,frome2013devise} and class annotations (e.g., attributes) \cite{lampert2013attribute}.
However, each kind of such external knowledge fails to accurately or fully express inter-class relationships.

Recently, Knowledge Graphs (KGs)~\cite{Pan2016,hogan2020knowledge,ehrlinger2016towards} %, which are a kind of graph structured knowledge, 
have attracted wide attention as the external knowledge of ZSL.
%, where structural knowledge is exploited.
For example, Wang et al. \cite{wang2018zero} and Kampffmeyer et al. \cite{kampffmeyer2019rethinking}
%\todo{[we don't normally use references as a subject, replace with authors or the name of the approach etc]}
%\cite{wang2018zero,kampffmeyer2019rethinking} 
incorporate hierarchical inter-class relationships from WordNet \cite{miller1995wordnet}; Gao et al. \cite{gao2019know}, Zhang et al. \cite{zhang2019integrating}, Nayak et al. \cite{nayak2020zero} and Roy et al. \cite{roy2020improving} explore relational class knowledge from common sense KGs such as ConceptNet \cite{speer2017conceptnet}.
Significant performance improvement is often achieved when these KGs are well utilized.
Besides, KGs can also be used to represent many other kinds of traditional external knowledge such as human annotations and textual information \cite{geng2021ontozsl,lee2018multi}, due to its high compatibility in representing and integrating different knowledge.
However, although various KGs have been exploited by current methods, there is still a concern on semantics completeness especially in distinguishing fine-grained classes.
Meanwhile, very few methods have been developed that can jointly utilize multiple kinds of knowledge in a KG, while other kinds of KG semantics such as logical expressions have not been investigated yet.
Furthermore, existing works all build their own KGs for evaluation, and the community lacks standard and unified benchmarks for comparing different KG-based ZSL methods under settings with ranging semantics.
And more importantly, KG-based external knowledge has been attractively investigated in the computer vision but rarely covered in other domains.
One critical reason for all these issues is that there is a shortage of high quality open benchmarking resources for method development and evaluation.

In this work, we constructed systemic resources for KG-based ZSL research. The resources include six benchmarks with corresponding KGs for three ZSL tasks from different domains: zero-shot image classification (ZS-IMGC), zero-shot relation extraction (ZS-RE), and zero-shot knowledge graph completion (ZS-KGC).
The KGs contain different kinds of external knowledge, including not only typical external knowledge such as attribute, text and hierarchy, but also relational facts and logical expressions, with the goal of providing ranging semantic settings for investigating different KG-based ZSL methods.
In the paper, we present the technical details of how these resources are constructed, their statistics, data formats, and very high usage for evaluating and developing robust and interpretable ZSL methods.

More importantly, we present an extensive benchmarking study by evaluating and comparing two representative and general ZSL methods, two setting-specific ZSL methods and one interpretable ZSL method, under different external knowledge settings that are supported by our resources.
%with these resources, including a state-of-the-art method that was originally developed to utilize KG and ontology external knowledge.
It is worth mentioning that there are currently no methods to utilize those potentially useful logical external knowledge, and 
%Moreover, to utilize these resources, various semantic embedding techniques have been utilized and developed. Especially, 
we thus developed an effective ensemble-based method which combines symbolic reasoning and neural prediction for ZS-KGC.
%to take advantage of the logical external knowledge from OWL to improve the completion results of ZS-KGC.
Through this benchmarking study, we analyzed the benefits of different external knowledge, and the pros and  cons of different ZSL methods. 
We have quite a few concrete observations and future work perspectives that will benefit the ZSL community and the KG community. See details in Section \ref{application} and \ref{sec:discussion}.
Here are two brief conclusions:
\begin{itemize}
    \item Utilizing various semantics represented by KGs can often lead to higher performance and more interpretable solutions, even when they are simply embedded and fed into some methods that are originally developed for single semantics. More effective methods for fusing and injecting different kinds of external knowledge should be investigated in the future. 
    \item ZSL methods of the generation-based paradigm often have more robust performance when both seen and unseen classes are predicted, than methods of the mapping-based paradigm, while the propagation-based paradigm can often well utilize the graph structure. In future ZSL studies, more ZSL methods of different paradigms should be tested and compared with, under different external knowledge settings. 
\end{itemize}
%Extensive experiment results show us the ability of KG-based external knowledge, the impact of different semantic settings, and the potential directions of KG-based ZSL researches, etc.
%We hope that our work will advance the zero-shot learning field especially the KG-based zero-shot learning field.

%\todo{we also demonstrated the high usability of these resources in evaluating the performance and explanation of two state-of-the-art KG-based ZSL methods and one classic ZSL method, and proposed a baseline method to utilize logical expressions in ZS-KGC.}

%\todo{[Add a list of bulletin points on our key contributions]}

The remainder of this paper is organized as follows.
In Section~\ref{background}, we set up the background of our work, including an introduction to KG and the ZSL tasks  in three different domains, review the related works.
%and give a close comparison of an old version of our work (the resources used in evaluating our method OntoZSL \cite{geng2021ontozsl}).
In the next three sections, we introduce the resources of ZS-IMGC, ZS-RE, and ZS-KGC, respectively.
In Section \ref{application}, we present the benchmarking study using these resources.
Subsequently, we summarize the evaluation results, and discuss the challenges and some potential research directions in Section~\ref{sec:discussion}.
In the end, we conclude the paper.

\section{Background}\label{background}
\subsection{Knowledge Graph}

%representation, integration and reasoning. 
Knowledge Graph (KG) is famous for representing and managing graph structured knowledge \cite{Pan2016,hogan2020knowledge,ehrlinger2016towards}. It has widely applied in many domains such as search engine, recommendation system, clinic AI, personal assistant, bioinformatics, intelligent finance, software engineering and data analysis \cite{PSAEZ2013,zou2020survey}. 
A KG is often largely composed of relational facts in the form of triples of Resource Description Framework (RDF)\footnote{ \url{https://www.w3.org/TR/rdf11-concepts/}.} \cite{Pan2009}.
Each RDF triple is denoted as ($s, r, o$), where $s$ represents a subject entity, $o$ represents an object entity, and $r$ represents a relation between these two entities (a.k.a. object property).
All these triples compose a multi-relational graph whose nodes correspond to entities and edges are labeled by relations.
A KG also contains RDF triples that represent literals and meta information such as entity attributes and textual definition via built-in or bespoke data and annotation properties such as \textit{rdfs:label} and \textit{rdfs:comment}.

In addition to these facts, KGs are often accompanied by an ontological schema and constraints in languages from the Semantic Web community such as RDF Schema (RDFS)\footnote{\url{https://www.w3.org/TR/rdf-schema/}}, Web Ontology Language (OWL)\footnote{\url{https://www.w3.org/TR/owl2-overview/}} and SHACL\footnote{\url{https://www.w3.org/TR/shacl/}} for richer semantics and higher quality  \cite{horrocks2008ontologies,baader2005description,hitzler2021review,paulheim2015serving}. They often defines entities' classes (a.k.a. concepts), properties (i.e., stating the terms used as relations), concept and relation hierarchies, constraints (e.g., relation domain and range, and class disjointness), and logical expressions such as relation composition.
The languages such as RDF, RDFS and OWL have defined a number of built-in vocabularies for representing these knowledge, such as \textit{rdfs:subClassOf}, \textit{rdf:type} and \textit{owl:disjointWith}. 
It is worth mentioning that a KG, especially those equipped with schemas and constraints, can support symbolic reasoning such as consistency checking, and entailment reasoning which infers hidden knowledge according to the defined logics. Some ontology reasoners such as HermiT \cite{DBLP:conf/owled/ShearerMH08} and TrOWL~\cite{TPR2010} can be directly applied. 

%Given a Knowledge Graph G = ($\mathcal{T}$, $\mathcal{A}$), where $\mathcal{T}$ is the schema sub-graph and $\mathcal{A}$ is the data sub-graph, we use the following semantic reasoning services:
%(1) Consistency checking: $\mathcal{G}$ is consistent, if there exists a model that satisfies all statements in $\mathcal{T}$ and $\mathcal{A}$.
%(2) Classification: this service computes all the subsumptions among named concepts in $\mathcal{T}$.
%(3) Entailment checking: this service checks if an axiom $\alpha$ is a consequence of $\mathcal{T}$ $\cup$ $\mathcal{A}$, or $\mathcal{T}$ $\cup$ $\mathcal{A}$ $\models \alpha$.
%(4) Materialisation: this service computes all individuals of concepts and roles in $\mathcal{G}$.  Reasoning services can be provided by existing reasoners,  such as  HermiT \cite{DBLP:conf/owled/ShearerMH08} and  TrOWL~\cite{TPR2010}.
%The expressiveness of knowledge graphs depends on the language used to express their schema. There are well known ways to simplify knowledge graph schema, such as approximations~\cite{PaTh07,RPZ2010,FMSP2012,PRZ2016,ZGNKH2015} and forgetting~\cite{WWTP2010,LuWo2011,WWTPA2014} to compute  reasoning results faithfully.  

Many kinds of data mining and machine learning techniques can also be applied to KGs for approximate inference and knowledge discover. 
One typical example is KG completion (KGC) by KG embedding techniques which are to learn vector representations of KG components such that their semantics such as relationships are kept in the vector space \cite{wang2017knowledge,LLSL+2015,yang2014embedding,gesese2019survey,chen2020owl2vec}.
KGC tasks often predict links between different KG components, such as between entities, between entities and classes, and between classes. Please see Section \ref{bg_kgc} for more details on KGC.
Another example is learning or mining concepts, rules, constraints and other ontological knowledge from KGs \cite{buhmann2016dl,zhang2019iteratively,volker2011statistical}. 
%Link prediction~\cite{LLSL+2015} can be used to produce an extension $\mathcal{A'}$ of the data sub-graph, of which the triples use only entities from $\mathcal{A}$ and  types / relations from $\mathcal{T}$. 
%One of the task of KGC is to predict the missing head $h$ or the missing tail $t$ of a triple. In this paper, we employ the knowledge graph embedding (KGE) approaches which embed entities and relations into a low-dimensional continuous vector space, so as to simplify operations on the KGs. The idea of embedding is to represent an entity as a k-dimensional vector \textbf{h} (or \textbf{t}) and defines a scoring function $f_r$(h,t) to measure the plausibility of the triple (h,r,t) in the embedding space. The representations of entities and relations are obtained by minimizing a global loss function involving all entities and relations. Different KGE algorithms often differ in their scoring functions, transformations, and loss functions.

%\todo{Literature on KGE, typical KGE methods, literal-aware KGE methods, multi-modal KGE, ontology embeddings, etc.}

\subsection{Zero-shot Image Classification}\label{bg_imgc}
Image classification is a critical task in computer vision.
Zero-shot image classification (ZS-IMGC) refers to predicting images with new classes that have no labeled training images.
In the literature of ZS-IMGC, case studies range from classifying general objects \cite{deng2009imagenet,farhadi2009describing} to classifying (fine-grained) objects in specific domains such as animals \cite{lampert2013attribute,xian2018zero}, birds \cite{welinder2010caltech}, and flowers \cite{nilsback2008automated}.
Please see \cite{xian2018zero} for a comprehensive survey on ZS-IMGC studies.

To address new classes, some early ZS-IMGC works employ class attributes as external knowledge, which describe objects' visual characteristics about e.g., colors and shapes, to model the relationships between classes.
However, these attributes ignore the direct associations between classes, cannot represent complicated relationship and usually need human labour for annotation.
Some other works adopt the word embeddings of class names \cite{frome2013devise,norouzi2013zero}, or the sentence embeddings or textual features of class descriptions \cite{reed2016learning} to model the inter-class relationships.
Although such textual information is easy to access, it cannot represent logical or quantitative semantics, and is often quite noisy containing many irrelevant words.

Recently, several methods model the inter-class relationships via KG, with promising results achieved.
Wang et al. \cite{wang2018zero} and Kampffmeyer et al. \cite{kampffmeyer2019rethinking} adopt WordNet to represent the hierarchy of classes of images from ImageNet;
Gao et al. \cite{gao2019know}, Nayak et al. \cite{nayak2020zero} and Roy et al. \cite{roy2020improving} propose to use common sense KG ConceptNet to introduce more relational knowledge;
Geng et al. \cite{geng2020explainable} extract knowledge from DBpedia as a complement of the WordNet class hierarchy.
However, all these KG-based ZS-IMGC studies are still preliminary in terms of both semantic sufficiency in the methodology and benchmarking in the evaluation.
To bridge the gap of benchmarking and support research in utilizing different external knowledge, in this work, we contributed three resources, each of which can support ranging external knowledge settings with a KG that has incorporated not only class hierarchy, text and attributes, but also common sense class knowledge and logical relationships between classes.

\subsection{Zero-shot Relation Extraction}\label{bg_re}

As an important semantic processing task in the field of natural language processing, the objective of
relation extraction (a.k.a. relation classification) is to predict the semantic relation of two given entity mentions in a sentence.
%by a specific sentence context.
Since the predicted relation and the given entity mentions can compose a relational fact, RE also serves as an essential technique for KG construction with text. 
Similar to image classification, conventional supervised relation extraction approaches cannot address new relation types that have never appeared in the training data.
To this end, the task of zero-shot relation extraction (ZS-RE), which is to predict unseen relations with given entity mentions and their sentences, was proposed and has been investigated by some studies \cite{LevySCZCoNLL2017,obamuyide2018zero,lockard2020zeroshotceres,imrattanatrai2019identifying,li2020logic,chen2021zs}.

To tackle these unseen relations, some ZS-RE studies convert the original problem to another text understanding problem by utilizing the relations' descriptive information.
%rely on a medium task to tackle these unseen relations.
For example, Levy et al. \cite{LevySCZCoNLL2017} reduce relation extraction to answering reading comprehension questions by associating one or more natural-language questions to each relation type.
Obamuyide et al.
\cite{obamuyide2018zero} formulate relation extraction as a textual entailment problem with the relation descriptions, and consider the input sentence and the description as the premise and hypothesis, respectively.
However, these works are labor intensive as human efforts are required to design questions or write descriptions for relations.
And the transformed tasks may not be suitable for the RE problem enough.
% \todo{And on the other hand, RE performance will be greatly affected by the ability of these medium models.}

Another attractive way is to leverage the external information that explicitly describes the semantic associations between relations, according to which unseen relations can be directly predicted by transferring features learned from seen relations.
For example, Chen et al. \cite{chen2021zs} explore them from the text descriptions of relation labels.
% \todo{For example, \cite{chen2021zs} leverage the descriptions of relations, i.e., text describing the relations and the associations between relations.}
% As the connection between RE task and downstream knowledge graph applications, 
Considering that relation labels can be represented in a KG by a number of relational facts,
the state-of-the-art is achieved by those who explored the semantics from KGs \cite{imrattanatrai2019identifying}.
One representative work is \cite{li2020logic} by our team, which builds associations between seen and unseen relations via implicit and explicit semantic representations with KG embeddings and logic rules.
In comparison with hand-crafted questions or  descriptions in aforementioned works, these external information contain more semantic knowledge about relations and are easier to collect such as accessing from online open resources.

Back to the external knowledge currently used, the structured knowledge by KGs is usually more accurate than the text, with less noise when incorporated in learning algorithms.
Thus, to better study the impact of KG external knowledge on ZS-RE  and facilitate developing more effective KG-based ZS-RE methods, we developed a new ZS-RE resource, where some data in \cite{li2020logic} are inherited and improved.
%provide a detailed introduction about how these resources are constructed in our paper.
In particular, targeting the situation that
the original benchmark used in \cite{li2020logic} is mainly for the standard zero-shot setting but ignores the more realistic generalized zero-shot setting,
% \todo{we make re-segmentation and down-sampling for the original training set, 
we segment the original training set,
and down-sample to extract two balanced subsets: one is taken as the new training set and the other is used for testing.
%Based on this, we comprehensively evaluate the different ZS-RE methods under settings with ranging external knowledge.
Besides, in the benchmarking study, we evaluated more ZS-RE methods such as OntoZSL \cite{geng2021ontozsl} under more settings, in comparison with the original paper \cite{li2020logic}.

\begin{table*}[]
\scriptsize
\centering
\renewcommand{\arraystretch}{1.3}
\caption{\small Comparison of our original resources used in OntoZSL \cite{geng2021ontozsl} and the new resources in KZSL (this paper).
}\label{tab:ontozsl_kzsl}
\begin{tabular}{c|ccccc|c|ccc}
\hline
\multirow{2}{*}{\begin{tabular}[c|]{@{}c@{}}Types of \\ External Knowledge\end{tabular}} & \multicolumn{5}{c|}{ZS-IMGC}       
& ZS-RE           
& \multicolumn{3}{c}{ZS-KGC} 
\\
% \cline{2-9}
& Hierarchies & Attributes & Literals & Relational Facts & Logics & KGs+Logic Rules & Text & RDFS    & OWL    
\\\hline
OntoZSL
& \Checkmark 
& \Checkmark  
& \Checkmark
& \XSolidBrush
& \XSolidBrush
& \XSolidBrush
& \Checkmark
& \Checkmark        
& \XSolidBrush
\\
% \hline
KZSL               
& \Checkmark 
& \Checkmark  
& \Checkmark
& \Checkmark
& \Checkmark
& \Checkmark
& \Checkmark
& \Checkmark        
& \Checkmark
\\\hline
\end{tabular}
\end{table*}

\subsection{Zero-shot Knowledge Graph Completion}\label{bg_kgc}
KGs such as Wikidata and DBpedia mostly face the challenge of incompleteness, and thus KG completion (KGC), which is often defined to predict the subject, relation or object of a missing triple (fact), has been widely investigated.
The KGC methods usually first embed entities and relations into vectors by e.g., geometric learning and Graph Neural Networks (GNNs), and then discover the missing facts in the vector space \cite{rossi2021knowledge}.
However, these methods can only predict entities and relations that have been associated in some training triples, but cannot address newly-added (unseen) entities and relations, which are quite common as KGs are often evolving.
To this end, zero-shot KGC (ZS-KGC), which is to predict triples with entities or relations that have never appeared in the training set, was recently proposed and has achieved quite much attention in recent years \cite{HamaguchiOSM2017OOKB,shah2019open,teru2019inductive,qin2020generative}.
As setting both entities and relations unseen makes the problem much more challenging, in this paper, we focus on those newly-added relations but keep the entity set seen.
We will consider unseen entities in the future.
\textbf{It is worth noting that ZS-RE and ZS-KGC here both involve completing relational facts with unseen relations, but they are totally different tasks.
ZS-RE aims to predict the missing relations given entity mentions and their associated text, while ZS-KGC aims to predict new facts given the existing relational facts.}

There are relatively few ZS-KGC studies that aim at addressing unseen relations.
Qin et al. \cite{qin2020generative} leveraged the features learned from relations' textual descriptions, and extracted two benchmarks from NELL and Wikidata for evaluation.
As in ZS-IMGC, textual external knowledge is usually noisy, with irrelevant words and ambiguous meanings.
To support further studies for developing and comparing ZS-KGC methods that can utilize different kinds of external knowledge,
we propose two ZS-KGC resources, each of which is associated with one KG composed of relational facts (i.e., \textbf{data graph}) as the target for completion (fact prediction), and one ontological schema (i.e., \textbf{schema graph}) as external knowledge.
For the schema graph, we adopt some vocabularies in RDFS, such as \textit{rdfs:domain}, \textit{rdfs:range}, \textit{rdfs:subPropertyOf} and \textit{rdfs:comment}, to define and describe relations with their e.g., domain and range constraints, hierarchy and text descriptions, and adopt some vocabularies in OWL, such as \textit{owl:inverseOf}, \textit{owl:propertyChainAxiom} and \textit{owl:SymmetricProperty}, to define some logics such as relation inversion and composition, and some characteristics of relations.

\subsection{Related Resources}
%With the deepening development of researches on ZSL, more and more studies have contributed their semantic resources to the community.
There have been some open resources that can be used for KG-based ZSL.
However, as already discussed above, the KGs of the existing resources usually have only one kind of semantics such as class hierarchy.
This makes it hard to fairly compare different methods that use different semantics, and limits the development of more effective methods that can fuse and utilize different KG semantics.
Meanwhile, the construction of these resources is usually very briefly introduced in evaluation sessions with details missing. This significantly limits their usage.
%However, most of them merely included the resource construction in the evaluation or the preliminary sections, and the construction details sometimes are not clear, which extremely limit the further study on them.
In contrast, our resources cover different tasks with KGs having different kinds of semantics, and the construction of these resources are well presented with details.
%to promote the knowledge-driven zero-shot learning, in this paper, we provide detailed illustrations on how the knowledge graph based semantic resources are constructed, from source data acquisition, graph-oriented data organization, to automatic filtering and crowd-assisted verification, so that contributing unified and standard benchmarks for the community.

A part of the proposed resources have already been very briefly introduced and used in our OntoZSL paper \cite{geng2021ontozsl} which focuses on presenting a new ZSL method.
However, these resources have been extended massively and some new resources have been added in this work.
As shown in Table \ref{tab:ontozsl_kzsl}, we \textit{i)} extended the KGs for ZS-IMGC with logical expressions and new common sense knowledge;
\textit{ii)} extended the ontological schemas for ZS-KGC with relation semantics in OWL;
\textit{(iii)} re-organized all the resources with formal knowledge representation, and higher accessibility;
and \textit{iv)} added a resource for a new task --- ZS-RE which is widely investigated in natural language processing.

It is worth noting that this is more than a resource paper, but includes an extensive benchmarking study.
We have evaluated  different ZSL methods using all these resources for different tasks, and have analyzed the impact of different kinds of semantics of the KGs.
Besides, we also present the use case of these resources for evaluating the explanations of some ZSL methods.

% Note an old version of these resources have been used in evaluating our method OntoZSL \cite{geng2021ontozsl}. This paper focuses on the resources, introducing their construction, statistics and use cases, and making them more accessible. The resources have also been substantially extended with logical expressions, new common sense knowledge, additional use cases, more formal KG representation and so on.

\section{Resource Construction for ZS-IMGC}\label{ZS_IMGC}
\subsection{Images and Classes}
We extract two benchmarks named ImNet-A and ImNet-O from ImageNet which is a large-scale image database organized according to the WordNet taxonomy \cite{deng2009imagenet}.
Each class in ImageNet is matched to a WordNet node, and has hundreds and thousands of images.
Due to a large number of hierarchical classes and a huge number of images, ImageNet is widely adopted in computer vision research as well as in ZSL research.

We focus on the class families (groups) in ImageNet, such as vehicles and dogs, and extract fine-grained classes with the following conditions:
1) seen classes are classes in the ImageNet 2012 1K subset that is often used to train CNNs e.g., ResNet \cite{he2016deep}, and unseen classes are those one-hop away according to the WordNet hierarchy;
2) the connection between seen and unseen classes are dense, e.g., a seen class has more than one neighboring unseen classes;
3) every class can be linked to a Wikipedia article such that more additional information about this class can be accessed;
and 4) the total number of selected seen and unseen classes in each family is  at least $5$.
As a result, we extracted $28$ seen classes and $52$ unseen classes from $11$ families all about animals (e.g., \textit{bees} and \textit{foxes}) for a benchmark named ImNet-A, and extracted $10$ seen classes and $25$ unseen classes for a benchmark named ImNet-O for general object classification from $5$ different class families (e.g., \textit{food} and \textit{fungi}).
Table~\ref{tab:imgc_datasets} shows detailed statistics of ImNet-A and ImNet-O.

In addition, we also re-use a very popular ZS-IMGC benchmark named Animals with Attributes (AwA) \cite{xian2018zero}.
AwA is a coarse-grained dataset for animal classification that contains $37, 322$ images from $50$ animal classes, all of which can be matched to WordNet nodes.
The original AwA benchmark has no KG, while in this work, we build a KG as its external knowledge.

\begin{table}
\scriptsize
\centering
\renewcommand{\arraystretch}{1.3}
\caption{\small Statistics of ZS-IMGC benchmarks. ``\#Att.'' refers to the number of attributes. S/U denote seen/unseen classes.
}\label{tab:imgc_datasets}
%\vspace{-0.3cm}
\begin{tabular}{c|p{0.6cm}<{\centering}|c|p{0.6cm}<{\centering}p{0.8cm}<{\centering}c}
\hline
\multirow{3}{*}{\textbf{Datasets}} 
% & \multirow{3}{*}{\bf Gran.}
&
\multirow{3}{*}{ \textbf{\#Att.}}
&
\multirow{2}{*}{\textbf{\#Classes}} & \multicolumn{3}{c}{ \textbf{\#Images}} 
\\
& & & 
& Training & Testing  \\ 
& 
& \multicolumn{1}{c|}{Total (S/U)}
&Total & S/U 
& \multicolumn{1}{c}{S/U}
\\
 \hline
ImNet-A 
% & fine
& 85 & 80 (28/52)
& 77,323 & 36,400/0 & 1,400/39,523
\\
ImNet-O 
% & fine 
& 40 & 35 (10/25)
& 39,361  & 12,907/0 
& \ \ \ 500/25,954
\\
AwA 
% & coarse
& 85 & 50 (40/10) 
& 37,322  & 23,527/0 & 5,882/7,913
\\
\hline
\end{tabular}
\end{table}

\subsection{External Knowledge and KG Construction}

We collect different kinds of external knowledge, including class hierarchy, attribute, text, relational fact and logical expression, for ImNet-A, ImNet-O and AwA. 
For each benchmark, we then construct one KG which integrates all these external knowledge.
The statistics of the resulting KGs are shown in Table~\ref{tab:imgc_resources}.

% class hierarchy
\subsubsection{Class Hierarchy}
\begin{figure*}
\centering
\includegraphics[width=0.8\textwidth]{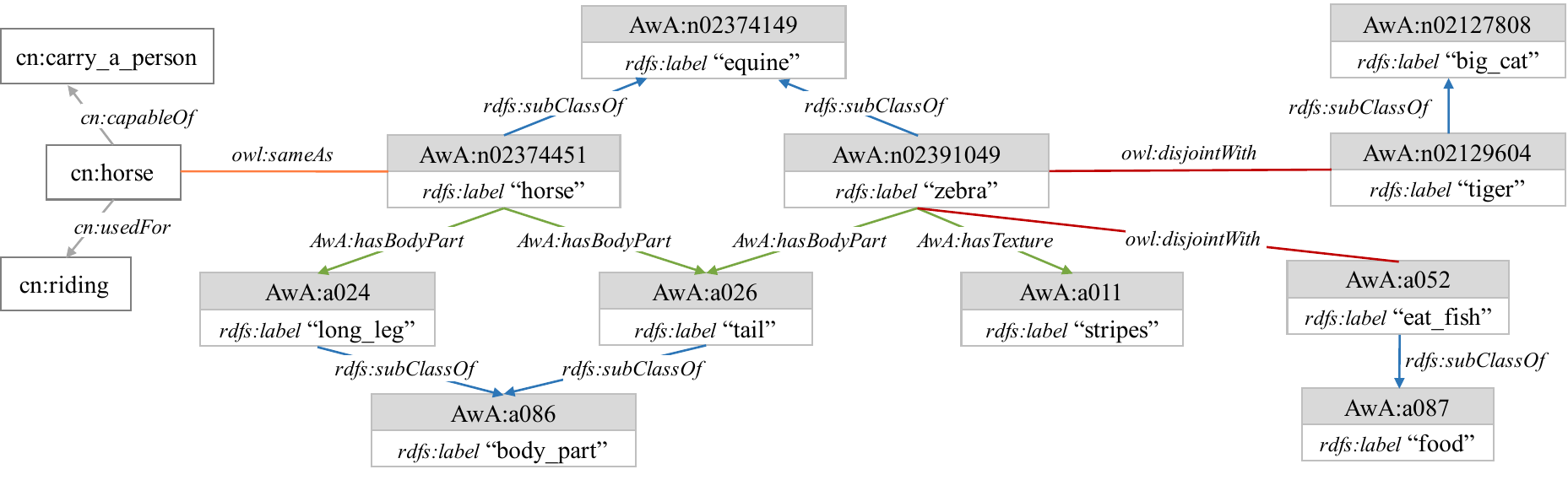}
%\vspace{-0.2cm}
\caption{\small A snapshot of the KG of AwA.
The prefixes \textit{AwA} and \textit{cn} are two ad-hoc namespaces of the KG.}\label{imgc_kg_example}
\end{figure*}

We first extract the class hierarchy from WordNet whose class nodes are connected via the \textit{super-subordinate} relation (a.k.a. hyponymy or hypernomy relation).
The class hierarchy is used as our KG backbone, and is formally represented by the RDFS vocabulary \textit{rdfs:subClassOf}, as Fig.~\ref{imgc_kg_example} shows.
Each class's IRI (Internationalized Resource Identifier) is created following its original WordNet id, e.g., \textit{zebra} from AwA has the IRI \textit{AwA:n02391049}.
The prefix ``\textit{AwA}'' here refers to an ad-hoc namespace of our KG.
Since WordNet contains a very large taxonomy, we extract a subset that covers all the benchmark classes and all their ancestors, using the WordNet interface in the Python package NLTK\footnote{\url{https://www.nltk.org/howto/wordnet.html}}.

\subsubsection{Class Attribute}
% class-attribute, attribute-attribute
Based on this structure, we then add the attribute annotations of seen and unseen classes to the graph.
Before adding, there is a need to establish the hierarchy of attributes.
This is because some attributes describe the same aspect of objects.
For example, attributes like \textit{black}, \textit{white} and \textit{red} all describe the appearance color of objects, while \textit{head}, \textit{tail} and \textit{claws} all describe the body parts of animals.
The categorization of attributes on the one hand models richer relationships among attributes, and on the other hand, it is helpful for defining the relations between classes and attributes.
Under the guidance of WordNet hierarchy, we manually gather the attributes in the datasets into different groups.
For example, we gather $17$ attributes into the group of body parts and $8$ attributes into the color group for the KG of AwA.

We represent these attributes and attribute groups as KG nodes, each of them also has a namespace specified by the dataset to which it belongs and a unique id defined by ourselves, as shown in Fig.~\ref{imgc_kg_example}.
Then, we connect attribute nodes to their group nodes via relation \textit{rdfs:subClassOf}, and define the relation edges from class nodes to attribute nodes according to the group to which the connected attribute belongs.
For example, for class \textit{zebra} and its one annotated attribute \textit{tail}, a relation named \textit{AwA:hasBodyPart} is defined.

Regarding the attribute annotation data, for AwA, we use its published class-attribute matrices annotated by experts \cite{lampert2013attribute}, in which each AwA class has an associated binary-valued or continuous-valued attribute vector.
In order to avoid the loss of semantic information,
% To represent the attribute annotation in the graph without losing information, 
we adopt the binary-valued version and extract attributes whose corresponding vector values are $1$ as annotated ones for each class.
While for ImNet-A/O, we manually annotate attributes for classes as the attributes of ImageNet classes are not available.
Briefly, we prepare a list of attributes that are gathered from Wikipedia pages and attribute annotations of other ZS-IMGC datasets such as AwA, divide all 115 classes into 5 parts, and invite 15 volunteers who are undergraduate from Zhejiang University for annotation.
Every volunteer is asked to assign 3$\sim$6 attributes for each class with the images and Wikipedia articles of classes as references.
Each class is independently reviewed by 3 volunteers and the final decision is made by voting.
The statistics of attributes of these three datasets are listed in Table~\ref{tab:imgc_datasets}.

\begin{table*}[]
\scriptsize
\centering
\renewcommand{\arraystretch}{1.3}
\caption{\small Statistics of different kinds of entities, relations and triples in the KGs of ZS-IMGC.
``Hie.'', ``Att.'' and ``CN'' are short for Hierarchy, Attribute and ConceptNet, respectively. 
}
 \label{tab:imgc_resources}
%\vspace{-0.3cm}
\begin{tabular}{c|cccc|ccc|ccccccc}
\hline
\multicolumn{1}{c|}{\multirow{2}{*}{\bf Datasets}} 
&
\multicolumn{4}{c|}{\bf \# Entities} 
& 
\multicolumn{3}{c|}{\bf \# Relations} 
&
\multicolumn{7}{c}{\bf \# Triples} 
\\ 
\multicolumn{1}{c|}{}
& {\bf Total} 
& Class & Att.
& CN
& {\bf Total} 
& Att. & CN
&
{\bf Total} 
& Hie. & Att.
& Literal & sameAs & CN &
% \todo{Logics}
disjointWith
\\
 \hline
ImNet-A 
& 8,920 & 111 & 103 & 8,706 
& 41 & 17 & 21
& 10,461 & 210 & 335 & 214 & 156 & 9,546 & /
\\
ImNet-O 
& 3,148 & 59 & 52 & 3,037
& 31 & 6 & 22
& 3,990 & 110 & 110 & 111 & 93 & 3,566 & /
\\
AwA 
& 9,195 & 100 & 102 & 8,993
& 42 & 15 & 23
& 14,112  & 197 & 1,562 & 202 & 182 & 10,546 & 1,423
\\
\hline
\end{tabular}
\end{table*}

\subsubsection{Class Text}
In addition to structured triples, we also introduce the textual information of classes and attributes.
Here, we choose their English surface names considering that some classes are hierarchically related and their names are similar.
For example, classes \textit{red\_fox}, \textit{grey\_fox}, \textit{kit\_fox} and their parent class \textit{fox}.
The class names can also be looked up by NLTK WordNet interface, we represent them in the graph using RDFS vocabulary \textit{rdfs:label}, as Fig.~\ref{imgc_kg_example} shows.

% External KGs.
\subsubsection{Relational Fact}
We also access more relational class knowledge from a large scale common sense KG ConceptNet \cite{speer2017conceptnet} whose knowledge is collected from multiple resources including WordNet, DBpedia, etc.
We use its latest dump\footnote{\url{https://github.com/commonsense/conceptnet5/wiki/Downloads}} and extract the English subset which contains over $3.4$ million triples and around $1.8$ million nodes in total.
It is obvious that not all of them contain valid information about the classes in the benchmarks, we therefore choose to extract a relevant subgraph by aligning classes and attributes to the entities of ConceptNet and querying their 1-hop neighbors.

Considering that entities in ConceptNet are words and phrases of natural language, we use the literal names of classes and attributes and conduct string matching for alignment.
For example, class \textit{zebra} can be aligned to ConceptNet entity \textit{c/en/zebra}.
For some attributes that cannot be matched due to different word forms, we lemmatize them before alignment. For example, attribute \textit{spots} is lemmatized to \textit{spot} that can be found in ConceptNet.
Besides, we also find that some ConceptNet entities refer to the same objects but have different forms, e.g., \textit{c/en/zebra} and \textit{c/en/zebra/n/wn/animal}.
Targeting this, we merge them using a custom namespace ``\textit{cn}'' and extract the union of these entities' neighborhoods.
For example, the above two entities are merged as \textit{cn:zebra}.
To be unified, other entities are also represented with this namespace.
Finally, for the aligned elements, we use a vocabulary \textit{owl:sameAs} defined in OWL to relate them in the graph.
From the statistics of resulting KGs shown in Table~\ref{tab:imgc_resources}, we find that the ConceptNet entities of some classes or attributes are still missing, it may be because they are fined-grained concepts and have not been included in ConceptNet yet.
We choose to skip them and leave the knowledge extraction of them as a future work.
In addition, to reduce the noise during neighborhood query, we ignore the relations with less information, e.g., Synonym, Antonym, SymbolOf, NotCapableOf and NotHasProperty.

However, to use the extracted subgraph, there are still some issues to be addressed.
One issue is that the relations extracted from ConceptNet may have the same semantics with the relations we have defined.
For example, \textit{cn:isA} and \textit{rdfs:subClassOf} both indicate the semantic of hierarchy.
For this, we unify them into \textit{rdfs:subClassOf}.
The other is that the knowledge extracted from ConceptNet may already exist.
An example is (\textit{cn:squirrel}, \textit{cn:LocatedNear}, \textit{cn:tree}), which is already modelled by the attribute triple: (\textit{AwA:squirrel}, \textit{AwA:hasHabitat}, \textit{AwA:tree}).
To solve this, we extract the subjects and objects of these triples to generate a set of tuples ($s, o$), and filter out ConceptNet triples with repetitive tuples.

\subsubsection{Logical Expression}
% OWL Logical Expressions
In ZS-IMGC, we also found that some classes that belong to different families and look greatly different have many identical attributes.
For example, two animal classes \textit{zebra} and \textit{tiger} both have attributes \textit{stripes}, \textit{tail} and \textit{muscle}.
During inference, \textit{tiger} may provide an unexpected significant contribution to the feature learning of \textit{zebra} due to too many shared attributes between them.
Although their parent classes (i.e., \textit{equine} and \textit{big\_cat}) and literal names have been introduced in the KG to distinguish them, more direct information would benefit the model and should be investigated.
One kind of semantics that can be expressed by a KG for further augmentation is the logical relationship defined using OWL vocabulary.
Therefore, we define disjointness for these classes and add disjointness axioms using built-in property \textit{owl:disjointWith}, as shown in Fig.~\ref{imgc_kg_example}.
In the example above, the disjointness between \textit{zebra} and \textit{tiger} means that an images of \textit{zebra} cannot simultaneously be the instance of \textit{tiger} so that avoiding the misclassification.
We also define the disjointness between classes and attributes.
For example, the fact ``Zebra doesn't eat fish'' means the disjointness between class \textit{zebra} and attribute \textit{eat\_fish}.

Since the overlap of attributes of ImNet-A/O classes in different families is low, we mainly set the disjoint constraints for classes and attributes in AwA.
For the disjointness between different classes, we first generate a candidate set by counting the number of shared attributes.
Specifically, for a pair of classes that belong to different families, if their shared attributes are more than $2/3$ of the attributes of class that has fewer attributes, we set a candidate disjoint relationship between them.
Then, we invite volunteers to check these candidates with their images as references so that ensuring the correctness of extracted class disjointness.
For the disjointness between classes and attributes, we leverage the continuous-valued attribute vectors, and each class is disjoint with attributes whose vector values are $0$.

\subsection{Data Overview and Storage}
In this section, we contributed two new fined-grained benchmarks ImNet-A and ImNet-O as well as their corresponding KGs.
ImageNet has been widely used in the ZS-IMGC literature due to its large scale and diverse granularity, while our work is the first attempt to extract and study its fined-grained subsets.
We also build a KG for the widely used coarse-grained dataset AwA.
Different from the external knowledge built in previous works with limited class semantics, our constructed KGs not only represent the widely used class knowledge --- class hierarchy, attributes and text in a unified graph, but also integrate a subgraph newly extracted from ConceptNet as well as some logic expressions that have not yet been investigated. 

% \todo{[Add a conclusion, providing some extensive discussions (with evidence) on illustrate how our work is better than existing benchmarks]}

Each constructed KG is composed of RDF triples which are stored in a CSV file with three columns corresponding to subjects, relations and objects.
The KG files can easily be accessed by Python libraries or be loaded into graph stores.
%such as RDFox \cite{nenov2015rdfox}.
Regarding the images, we follow previous work \cite{xian2018zero} and provide ResNet features which are stored as a matrix, whose two dimensions  correspond to feature vector length and image number, respectively.

\section{Resource Construction for ZS-RE}

\subsection{Relation Text and Types}
In the previous work \cite{li2020logic}, we contributed a benchmark for ZS-RE which is constructed from a well-known relation extraction dataset named Wikipedia-Wikidata \cite{sorokin2017context}.
Due to the tight integration of Wikipedia and Wikidata, Wikipedia-Wikidata collects sentences from English Wikipedia corpus, identifies Wikidata entities in the sentences, and annotates relation labels by querying Wikidata relations that connect the extracted entities.
As a result, a number of samples are generated, each of them is accompanied by a sentence text, a pair of entity mentions and a relation between them.

%To evaluate the proposed ZS-RE method, 
To construct the ZS-RE benchmark, all relations in the Wikipedia-Wikidata are first clustered based on their word embeddings, and then the relations in each cluster are divided into two disjoint groups --- the seen group and the unseen group according to the number of their samples.
The first step ensures the preliminary semantic associations (from word embeddings) between seen and unseen relations, while the second step ensures that seen relations are data-rich while unseen relations are data-poor.
In \cite{li2020logic}, the relations with more than 1,200 samples are specified as seen relations while the rest are specified as unseen ones. 
% \todo{
% In the paper, the authors specify relations with more than $1200$ instances as seen ones while the rest as unseen relations.
% After dropping unseen relations whose instance numbers are less than $500$ from the cluster and manually adjusting}, 
After manual adjustments, a dataset with $70$ seen relations and $30$ unseen relations is finally outputted.

However, the dataset focuses on serving the standard zero-shot setting where samples of only unseen relations are tested, while ignores a more realistic generalized ZSL (GZSL) setting where samples of seen and unseen relations both appear during testing.
%, and the candidate label space contains both seen and unseen relations (i.e., generalized ZSL, GZSL).
Moreover, the training samples are not well balanced with respect to the relation --- 44 out of 70 seen relations all have 2,800 samples, while the rest 26 have less than 2,800 samples (the minimum is 1,246).
The imbalanced training set may have a negative impact on the quality of pre-training a deep neural network that will be used to extract the features of samples.
% the maximum used for training is $2,800$ while the minimum is $1,246$.
To support the GZSL setting and release the imbalance issue, in this paper, we further split the current training set.
Firstly, we down-sample the training set to 1,200 instances per relation to make the training samples balanced over different seen relations.
Then, we further down-sample the rest data to the maximum of 50 sentences per relation to generate another balanced subset as the testing data of seen relations.
The sample numbers of the original dataset from \cite{li2020logic} and the new dataset are compared as Table~\ref{tab:re_datasets} shows.
We rename the new dataset as \textbf{ZeroRel}.

%and released it for supporting subsequent studies on ZS-RE. 

\begin{table}
\scriptsize
\centering
\renewcommand{\arraystretch}{1.3}
\caption{\small Statistics of the ZS-RE benchmarks. S/U means seen/unseen relations.
}\label{tab:re_datasets}
%\vspace{-0.3cm}
\begin{tabular}{c|c|p{0.8cm}<{\centering}p{0.9cm}<{\centering}c}
\hline
% \multirow{2}{*}{\bf Datasets} 
% & 
% & Training & Testing  \\ 
% &Total & S/U 
% & \multicolumn{1}{c}{S/U}
% \\
\multirow{3}{*}{\bf Datasets} 
&
\multirow{2}{*}{\bf \# Relations } & \multicolumn{3}{c}{\bf \# Sentences} 
\\
& &
& Training & Testing  \\ 
& \multicolumn{1}{c|}{S/U}
&Total & S/U 
& \multicolumn{1}{c}{S/U}
\\
 \hline
Origin Data \cite{li2020logic}
& 70/30
& 193,867 
& 176,717/0 
& \quad \quad 0/17,150
\\
ZeroRel (New)
& 70/30
& 104,646  & 84,000/0
& 3,496/17,150
\\
\hline
\end{tabular}
\end{table}

\subsection{Knowledge Graph and Logic Rule}\label{zsre_semantics}

%We inherit the external knowledge resources contributed by \cite{li2020logic} and introduce how them are constructed in detail, including a knowledge graph and some logic rules.

We next present how the KG and rule external knowledge are constructured for ZeroRel.
%\subsubsection{Knowledge Graph}
%
%In \cite{li2020logic}, the authors first
As introduced in \cite{li2020logic}, the implicit semantic association between seen and unseen relations is first mined from pre-trained KG embeddings following the assumption that the embeddings of semantically similar relations are located to each other in the embedding space \cite{yang2014embedding}.
For example, the similarity of vectors of relations \textit{nationality} and \textit{live\_in\_country} is higher, whereas the relation \textit{profession} has low similarity with the former two relations.
Therefore, given Wikidata entity mentions and relations in the dataset, we directly use the Wikidata knowledge graph and adopt a version of dump\footnote{\url{http://openke.thunlp.org/download/wikidata}} that is often used for training KGE models.
Statistically, the dump contains 20,982,733 entities, 594 relations and 68,904,773 triples in total.
For convenience, we do not set additional namespaces for it.
Based on such a KG, various KGE methods can be applied to pre-train semantically meaningful representations for relation labels and build the implicit semantic associations.
From another point of view, the semantic associations between relations can also be modeled by this KG in a symbolic way, e.g., a batch of shared neighboring entities.

% \todo{to obtain textual descriptions of KG elements, we extract a subset from it.}
% how to extract and statistics.

%\subsubsection{Logic Rule}\label{logic_rules}
Logic rules are further constructed to represent the associations between relations.
They are in the form of \textit{body} $\Rightarrow$ \textit{head}, where \textit{head} is a binary relation and \textit{body} is a conjunction of binary and unary relations, and the length of a rule is determined by the number of relations in its \textit{body}.
A rule of length $2$ is like $\textit{brother} (x, y) \wedge \textit{father} (y, z) \Rightarrow \textit{uncle} (x, z)$, where $x, y, z$ are three entity variables, showing the instances of relation \textit{uncle} can be inferred by the instances of relations \textit{brother} and \textit{father}, and the relation \textit{uncle} can be viewed as a composition of relations \textit{brother} and \textit{father}.
A rule of length $1$ such as $\textit{born\_in\_country}(x, y) \Rightarrow \textit{nationality}(x, y)$ illustrates the semantic identity between relations \textit{born\_in\_country} and \textit{nationality}.
In the ZSL setting, if a rule involves an unseen relation, its instances can be predicted based on other seen relations in the rule.
Therefore, it is intuitive to incorporate with logic rules to build an explicit association between seen and unseen relations.

Logic rules could be automatically extracted from structured KGs by any KG rule mining algorithms or tools.
We choose AMIE \cite{galarraga2013amie} for its convenience and fast-speed to extract logic rules of different lengths from the Wikidata KG proposed above.
Each mined rule is associated with a PCA confidence score provided by AMIE.
In particular, we limit the maximum length of rules to 2 for the efficiency of mining valid rules, and then keep those rules which include at least one relation in the dataset with a confidence threshold of $0.3$.
Finally, we mined 50 length-1 rules and 122 length-2 rules in total for the relations in the dataset.
\textbf{It is noted that these rules can also be formulated using OWL axioms but with probability values, i.e., the length-1 rules with higher confidence values have higher possibility to mean the relation equivalence that can be represented by OWL vocabulary \textit{owl:equivalentProperty}, and the length-2 rules with higher confidence values have higher possibility to mean the relation composition.}

\subsection{Data Overview and Storage}
To evaluate the zero-shot learning in relation extraction, we constructed a ZS-RE benchmark and contributed a new version that supports more ZSL settings.
Based on this, we contributed two kinds of external knowledge --- KG and logic rules, both of them contain richer and more accurate relation semantics than the base one from word embeddings.

% \todo{[Add a conclusion, providing some extensive discussions (with evidence) on illustrate how our work is better than existing benchmarks]}

We store the KG resource in a CSV file as ZS-IMGC case, with three columns corresponding to the subject entities, relations and object entities.
The extracted logic rules are stored in a JSON file with ``head'', ``body'' and ``pcaconf'' properties specifying the \textit{head}, \textit{body} and PCA confidence score of a rule.
As for the ZS-RE data ZeroRel, we follow previous work to provide it in a CSV file, in which each row corresponds to a sample including the sentence text, the relation label, the entity mention pairs and their indexes in the sentence.

\section{Resource Construction for ZS-KGC}\label{ZS_KGC}

\subsection{KGs for Completion}
We employ two ZS-KGC datasets (i.e., two sub-KGs for completion) proposed in \cite{qin2020generative}.
They are NELL-ZS extracted from NELL\footnote{\url{http://rtw.ml.cmu.edu/rtw/}} and Wiki-ZS extracted from Wikidata\footnote{\url{https://www.wikidata.org/}}.
In both datasets, relations are divided into two disjoint sets: a seen relation set $\mathcal{R}_s$ and an unseen relation set $\mathcal{R}_u$.
In training, a set of facts (RDF triples) of the seen relations are used, while in testing, the model predicts the facts involving unseen relations in $\mathcal{R}_u$.
A closed set of entities are considered in both datasets, which means each entity in the testing triples has already appeared in the training triples.
Besides, a subset of training facts is left out as the validation set by filtering all training facts of the validation relations. 
The statistics of these two datasets are shown in Table~\ref{tab:kgc_datasets}.

\begin{table}
\scriptsize
\centering
\renewcommand{\arraystretch}{1.3}
\caption{\small Statistics of ZS-KGC datasets. ``Tr/V/Te'' is short for training/validation/testing.
``\#Ent.'' denotes the number of entities.
}
 \label{tab:kgc_datasets}
%\vspace{-0.3cm}
% \begin{tabular}{c|p{0.8cm}<{\centering}|p{2cm}<{\centering}|p{3.7cm}<{\centering}}
\begin{tabular}{c|c|c|c}
\hline
\multicolumn{1}{c|}{\multirow{2}{*}{\bf Datasets}} 
 &
\multicolumn{1}{c|}{\multirow{2}{*}{\bf \# Ent.}}
&
\multicolumn{1}{c|}{\bf \# Relations} 
& \multicolumn{1}{c}{\bf \# Triples} 
\\

& 
& \multicolumn{1}{c|}{Total (Tr/V/Te)}
&Total (Tr/V/Te)
\\
 \hline
NELL-ZS 
& 65,567 & 181 (139/10/32) 
& 188,392 (181,053/1,856/5,483) 
\\
Wiki-ZS 
& 605,812 & 537 (469/20/48) 
& 724,928 (701,977/7,241/15710) 
\\
\hline
\end{tabular}
\end{table}

\subsection{Ontological Schema}
We build an ontological schema as external knowledge for each dataset. It mainly includes semantics expressed by RDFS (concept hierarchy, relation hierarchy, relation's domain and range), semantics expressed by OWL (relation characteristics and inter-relation relationships), and textual meta data (e.g.,  the names and descriptions of concepts and relations).
Note concept here refers to entity type/class.
In our paper, the sub-KG for completion is also named as a data graph, and its ontology schema is a schema graph.
The statistics of the resulting ontological schemas of NELL-ZS and Wiki-ZS are shown in Table~\ref{tab:kgc_resources}.

\subsubsection{Semantics in RDFS}
Semantics by RDFS vocabularies act as the backbone of the schema graph.
Different from the data graph where relations act as edges between nodes, in the schema graph, relations act as nodes (i.e., subjects or objects in RDF triples).
Specifically, we use the vocabularies \textit{rdfs:subPropertyOf}, \textit{rdfs:domain}, \textit{rdfs:range} and \textit{rdfs:subClassOf} to define the relation semantics and generate corresponding triples:
\begin{itemize}[leftmargin=0.4cm]
    \item ($r_1$, \textit{rdfs:subPropertyOf}, $r_2$), subproperty triple, states the hierarchical relationships between relations, i.e., relation $r_1$ is a subrelation of relation $r_2$;
    \item ($r$, \textit{rdfs:domain}, $C_s$), domain triple, summarizes the subject entity type (i.e., subject concept) $C_s$ of relation $r$;
    \item ($r$, \textit{rdfs:range}, $C_o$), range triple, summarizes the object entity type (i.e., object concept) $C_o$ of relation $r$;
    \item ($C_i$, \textit{rdfs:subClassOf}, $C_j$), subclass triple, states the hierarchical relationships between entity types $C_i$ and $C_j$.
\end{itemize}

\begin{figure*}
\centering
\includegraphics[width=0.9\textwidth]{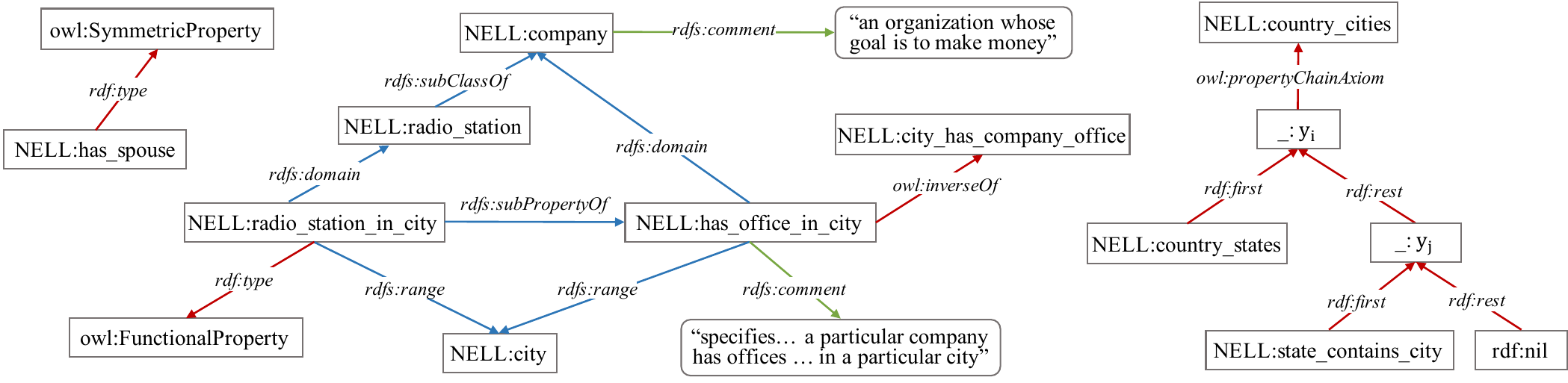}
\vspace{-0.2cm}
\caption{\small A snapshot of the constructed schema graph for ZS-KGC dataset NELL-ZS.
%\todo{The prefixes+illustration for the right part.}
\_:$y_i$ and \_:$y_j$ denote two blank nodes.
} \label{kgc_kg_example}
\end{figure*}

A snapshot of the schema graph of NELL-ZS is shown in Fig.~\ref{kgc_kg_example}.
As we can see, the schema graph's nodes are relations and entity types of the data graph.
We also add dataset-specific namespace for these nodes.
Besides, the edge in the schema graph, i.e., the relationship between relations, is called as ``meta-relation''.

For the schema graph of NELL-ZS, the aforementioned semantics in RDFS can be extracted from NELL's ontology.
The ontology is saved and published as a CSV file\footnote{\url{http://rtw.ml.cmu.edu/resources/results/08m/NELL.08m.1115.ontology.csv.gz}} which has three columns corresponding to subjects, predicates and objects of RDF triples.
From these triples, we extract domain and range triples according to the predicates ``domain'' and ``range'', respectively, and extract subproperty and subclass triples according to the predicate ``generalizations''.
For the schema graph of Wiki-ZS, these semantics can be accessed from Wikidata by a tookit implemented in Python\footnote{\url{https://pypi.org/project/Wikidata/}}.
Specifically, we look up a relation's super-relations by Wikidata property P1647 (subproperty of), look up a relation's domain concepts and range concepts by Wikidata property P2302 (property constraint) with constraints Q21503250 (type constraint) and Q21510865 (value-type constraint), respectively,
and look up a concept's super-concepts by Wikidata property P279 (subclass of).

% Literals
\subsubsection{Semantics in Text}
We further enrich the schema graphs with textual information of the nodes (i.e., relations and concepts), which usually act as important external knowledge in addressing ZS-KGC \cite{qin2020generative,geng2021ontozsl,shah2019open}.
For NELL-ZS, we extract the textual descriptions of relations and concepts from NELL's ontology file by the predicate ``description''.
For Wiki-ZS, we look up the surface names and descriptions of relations and concepts from Wikidata using properties \textit{label} and \textit{description}, respectively.
The extracted text can be represented in the graph by RDFS vocabularies \textit{rdfs:label} and \textit{rdfs:comment}, leading to a literal-aware schema graph.

\begin{table*}[]
\scriptsize
\renewcommand{\arraystretch}{1.3}
\caption{\small Illustrations and statistics of inter-relation relationships and relation characteristics of the ontological schemas of NELL-ZS and Wiki-ZS.
$x, y, z$ are entity variables.
``[NELL]'' and ``[Wiki]'' denote the example comes from NELL-ZS and Wiki-ZS, respectively.}
 \label{tab:owl_illustrations}
%\vspace{-0.3cm}
\begin{tabular}{llllcc}
% \hline
\toprule[0.6pt]
\multicolumn{1}{c}{\multirow{2}{*}{\begin{tabular}[c]{@{}c@{}}OWL\\ Semantics\end{tabular}}}
& \multicolumn{1}{c}{\multirow{2}{*}{Formula}} 
& \multicolumn{1}{c}{\multirow{2}{*}{Example}}  
% & Triple Form
& \multicolumn{2}{c}{Statistics} \\
& \multicolumn{1}{c}{}                  
& 
% & 
& NELL-ZS  & Wiki-ZS \\
\midrule[0.4pt]
Inversion     
& $(x, r_1, y) \Leftrightarrow (y, r_2, x)$
% & childOf \& parentOf 
& P802 (student) \& P1066 (student of) [Wiki]
% & (P2354, \textit{owl:inverseOf}, P360)
& 0 & 39
\\
% \todo{Equivalence}   
% & $r_1(x,y) \Leftrightarrow r_2(x,y)$
% & cost \& price 
% & 0 & 0             
% \\
% Disjointness  % Mutual Exclusion
% & $(?x, r_1, ?y) \wedge (?x, r_2, ?y) \Rightarrow \bot$ 
% % & fatherOf \& motherOf 
% & \todo{sportfansincountry \& sportschoolincountry [NELL]} 
% & 4 & 6   
% \\
Composition   
& $(x, r_1, y) \wedge (y, r_2, z) \Rightarrow (x, r_3, z)$
& countrystates $\wedge$ statecontainscity $\Rightarrow$ countrycities [NELL]
% & brotherOf $\wedge$ parentOf $\Rightarrow$ uncleOf
& 20
& 7
\\
% \hline
\midrule[0.4pt]
Symmetry
& $(x, r, y) \Rightarrow (y, r, x)$
% & hasSibling
& hasspouse [NELL]
% & (\textit{hasspouse}, \textit{owl:SymmetricProperty}, \textit{true})
& 20 & 25 
\\
Asymmetry 
& $(x, r, y) \Rightarrow \neg (y, r, x)$
% & sonOf
& subpartof [NELL]
% & (\textit{subpartof}, \textit{owl:AsymmetricProperty}, \textit{true})
& 24 & 11
\\
Reflexivity   
& $(x, r, x)$
% & knows  
& animalpreyson [NELL]
% & (\textit{animalpreyson}, \textit{owl:ReflexiveProperty}, \textit{true})
& 2               & 0
\\
Irreflexivity 
& $\neg (x, r, x)$
% & marriedTo
& P184 (doctoral advisor) [Wiki]
% & (P184, \textit{owl:IrreflexiveProperty}, \textit{true})
& 46 & 15       
\\
Functionality
& $(x, r, y) \wedge (x, r, z) \Rightarrow y=z$
% & hasBirthMother
& airportincity [NELL]
% & (\textit{airportincity}, \textit{owl:FunctionalProperty}, \textit{true})
& 6 & 15  
\\
% \multicolumn{1}{l}{\begin{tabular}[c]{@{}c@{}}Inverse\\ Functional\end{tabular}}
Inverse Functionality
& $(x, r, y) \wedge (z, r, y) \Rightarrow x=z$
% & isBirthMotherOf
& statecontainscity [NELL]
% & (\textit{statecontainscity}, \textit{owl:InverseFunctionalProperty}, \textit{true})
& 16
& 1
\\
% \todo{Transitivity}
% & $r_1(x,y) \wedge r_1(y,z) \Rightarrow r_1(x,z)$
% & ancestorOf   
% & 0 & 0  
% \\
\bottomrule[0.6pt]
\end{tabular}
\end{table*}

\subsubsection{Semantics in OWL}\label{OWL_Axioms}
We also introduce relation semantics in OWL including the relationships between relations and relation characteristics.
We provide an overview illustration with definitions and examples in Table~\ref{tab:owl_illustrations}, and will next introduce the details.
%Next, we will introduce one by one.

\textbf{Inverse Relationship.}
The inverse relationship between two
relations is defined by \textit{owl:inverseOf}.
If $r_1$ is an inverse relation of $r_2$, when a fact $(e_1, r_1, e_2)$ holds, the fact $(e_2, r_2, e_1)$ also holds, and vice versa.
In building the ontological schemas for ZS-KGC, we introduce the inverse relationships between seen and unseen relations, with triples in format of ($r_1$, \textit{owl:inverseOf}, $r_2$). 
% In predicting triples with unseen relations, 
In prediction, the ZS-KGC models can utilize the unseen relations' inverse relations which have been involved in the training triples.
Since the inverse relations have been removed from NELL-ZS when it is originally constructed, we only add inverse triples for relations in Wiki-ZS, which are extracted from Wikidata by its property P1696 (inverse property).

\textbf{Compositional Relationship.}
%As we introduced in Section~\ref{logic_rules}, the logic rules state that 
A relation can be constructed by ordered composition of several other relations.
Relation $r_3$ is a composition of another two relations $r_1$ and $r_2$, denoted as $r_1  \wedge r_2 \Rightarrow r_3$, if we have $(x, r_1, y) \wedge (y, r_2, z) \Rightarrow (x, r_3, z)$, where  $x$, $y$ and $z$ are three entity variables.
Such compositional relationships are also helpful for KGC.
For example, with $\textit{brother} \wedge \textit{father} \Rightarrow \textit{uncle}$, we can infer $a$ is an uncle of $c$ if $a$ is a brother of $b$ and $b$ is a parent of $c$.
Therefore, we add composition axioms in our ontological schemas to define some seen and unseen relations as the compositions of some seen relations.
We limit the number of compositional relations in each axiom to 2.
%and formally define the semantics of relation composition as
%\begin{equation}\label{eq:composition}
%    (x, r_1, y) \wedge (y, r_2, z) \Rightarrow (x, r_3, z)
%\end{equation} $r_1  \wedge r_2 \Rightarrow r_3$ in short, where $r_1$, $r_2$ and $r_3$ denote three relations, and $x$, $y$ and $z$ are three  variables of entities.
A composition axiom can be represented in the schema graph as the rightest part of Fig. \ref{kgc_kg_example} shows, where it has been serized as RDF triples with blank nodes according to W3C OWL to RDF graph mapping standard\footnote{\url{https://www.w3.org/TR/owl2-mapping-to-rdf/}}.

For NELL-ZS, we first extract a set of candidate relation compositions via checking relation's domain and range.
Specifically, for any three relations $r_1$, $r_2$ and $r_3$, if the range of $r_1$ is the domain of $r_2$, the domain of $r_1$ is also the domain of $r_3$ and the range of $r_2$ is also the range of $r_3$, then $r_1  \wedge r_2 \Rightarrow r_3$ is regarded as a candidate composition.
These candidates can be extracted according to the schema of NELL-ZS defined by RDFS. 
Briefly, for each relation of NELL-ZS, we traverse all seen relation pairs and check whether the domains and ranges of the two seen relations and the current relation match the above condition.
Some candidate relation compositions extracted in the above step are not correct; one example is $ \textit{mother\_of\_person} \wedge \textit{person\_also\_knownas} \Rightarrow \textit{wife\_of}$.
%does not match common sense.
Targeting this, we manually check these candidates.
Briefly, each candidate is independently reviewed by three volunteers (including one of the authors and two of our colleagues who are familiar with KGs and ontologies) and the final decision is made by voting.
It is also allowed that volunteers can look up all the information about relations such as triples and descriptions during review.

However, the method of checking relation's domain and range can not be well applied to Wiki-ZS because most Wiki-ZS relations have multiple domains and multiple ranges, which will result in too many candidates and cost too much manual assessment.
Therefore, for Wiki-ZS, we use AMIE to mine compositional rules from facts.
Different from mining rules from the Wikidata dump in the ZS-RE task, we mine relation compositions from the triples of Wiki-ZS dataset since some Wiki-ZS relations are not included in the dump.
Moreover, to ensure the correctness of mined rules, we \textit{i)} filter out those rules whose scores are below $0.9$, and \textit{ii)} invite volunteers to manually assess the remaining rules as for NELL-ZS.
Finally, we transform the correct ones into relation composition axioms for the schema of Wiki-ZS.

\begin{table*}[]
\scriptsize
\centering
\renewcommand{\arraystretch}{1.3}
%\vspace{-0.2cm}
\caption{\small Number of relations, concepts, literals, meta-relations, and different axioms in the ontological schema.
}
 \label{tab:kgc_resources}
\begin{tabular}{c|cccc|ccccccccc}
\hline
\multicolumn{1}{c|}{\multirow{1}{*}{\bf Datasets}} 
% &
% \multicolumn{3}{c|}{\bf \# Concepts} 
% & 
% \multicolumn{3}{c|}{\bf \# Meta-relations} 
% &
% \multicolumn{7}{c}{\bf \# Triples} 
% \\ 
% \multicolumn{1}{c|}{}
& \# relations & \# concepts &  \# literals &  \# meta-relations
&  \# subproperty &  \# domain &  \# range &  \# subclass  &  
\multicolumn{1}{c}{\begin{tabular}[c]{@{}c@{}} \# relation \\ characteristics \end{tabular}}
\\
 \hline
NELL-ZS
& 894 & 292 & 1,063 & 9 & 935 & 894 & 894 & 332  & 114

\\
Wiki-ZS 
& 560 & 1,344 & 3,808 & 11
& 208 & 1,843 & 1,378  & 1,392  & 67
\\
\hline
\end{tabular}
\end{table*}

\textbf{Symmetry \& Asymmetry.}
In a KG, a relation $r$ is symmetric if we have $(y, r, x)$ given $(x, r, y)$. 
One typical example is \textit{has\_spouse}.
In contrast, a relation $r$ is defined as asymmetric if $(y, r, x)$ is always false given $(x, r, y)$.
We add symmetric and asymmetric characteristics to some relations in our schemas of NELL-ZS and Wiki-ZS, because they could be utilized by potential methods for addressing ZS-KGC  by e.g., inferring more facts for training and finding similar seen relations for an unseen relation.
%For example, when an unseen relation is symmetric, during prediction, some of its  testing triples could be inferred according to some other triples that have been correctly predicted.
%The asymmetric property could reduce the searching space of candidate entities during prediction.
%Namely, entity $e_1$ can be directly excluded from being the object of the tuple ($e_2, r$) without predicting if the triple $(e_1, r, e_2)$ holds.
In our resource we add symmetry and asymmetry for relations that have identical domains and ranges.

To add symmetry and asymmetry for relations in NELL-ZS, we use the predicate ``anti-symmetric'' defined in the ontology file of NELL.
Specifically, for symmetric relations, we first extract relations whose ``anti-symmetric'' values are false, and then select those with the same domain and range.
Some of the resultant relations are still not symmetric, and we invite volunteers to filter out them.
For asymmetric relations, they can be automatically extracted by the predicates ``anti-symmetric'' and ``irreflexive'' considering that a relation is asymmetric iff it is antisymmetric and irreflexive.
For relations of Wiki-ZS, the symmetric relations can be extracted by looking up the Wikidata constraint Q21510862 (symmetric constraint) stated in the property P2302 (property constraint).
While for the asymmetry, we extract relations which have identical domain and range, and manually assess them.
%whether the relation is asymmetric or not.

We use membership axioms and two OWL built-in concepts \textit{owl:SymmetricProperty} and \textit{owl:AsymmetricProperty} to represent relation symmetry and asymmetry in our ontological schemas.
When the ontologies are represented as schema graphs, relation characteristics are transformed into RDF triples like ($r, \textit{rdf:type}, \textit{owl:SymmetricProperty}$) which means $r$ is a symmetric relation.

\textbf{Reflexivity \& Irreflexivity.}
A relation $r$ is regarded as reflexive if $(x, r, x)$ holds and as irreflexive if $(x, r, x)$ does not hold where $x$ is an entity variable.
Similar to symmetry and asymmetry, relation reflexivity and irreflexivity could be utilized in ZS-KGC with e.g., additional training samples and more information for relation similarity.
We can even directly infer testing triples in form of $(e, r, e)$ if $r$ is reflexive.
%if $r$ is irreflexive, we can exclude entity $e$ as the object of the tuple ($e, r$) in prediction.

We use the values of the predicate ``anti-reflexive'' defined in the NELL's ontology file to add reflexive and irreflexive characteristics for relations in NELL-ZS.
%i.e., relations annotated with false are reflexive while those annotated with true are irreflexive.
For relations in Wiki-ZS, since Wikidata has no definitions towards these two characteristics,
we extract relations that have identical domain and range, and manually assess their reflexivity and irreflexivity.
The representation of relation reflexivity and irreflexivity in the schema graph is the same as symmetry and asymmetry but uses the built-in concepts of \textit{owl:ReflexiveProperty} and \textit{owl:IrreflexiveProperty}.

\textbf{Functionality \& Inverse Functionality.}
Given a functional relation $r$, if $(x, r, y)$ and $(x, r, z)$ holds, then $y$ and $z$ must be the same entity.
Namely every entity can be related to at most one entity via a functional relation.
A relation can also be defined as inverse functional when its inverse relation is functional.
We add both functionality and inverse functionality to some relations.
They can be potentially used for addressing ZS-KGC as the other relation characteristics discussed above.
%in our ontologies.
They can also constrain the searching space for new triples. For example, given a subject and a functional relation, the corresponding object must be unique.
%($e, r$) with functional relation $r$, its ground truth object entity is unique.

To extract these two characteristics, we look up the whole set of triples of NELL (via its published dump\footnote{\url{http://rtw.ml.cmu.edu/resources/results/08m/NELL.08m.1115.esv.csv.gz}}) and Wikidata (via its SPARQL Endpoint), and extract relations whose object entity is unique for the same subject, and relations whose subject entity is unique for the same object.
%for the inverse functional characteristic.
They are represented in the same way as symmetry and asymmetry but use the built-in concepts of \textit{owl:FunctionalProperty} and \textit{owl:InverseFunctionalProperty}.

%More formulated illustrations of these relation semantics and their examples are provided in Table~\ref{tab:owl_illustrations}.
%Table~\ref{tab:kgc_resources} shows the statistics of the resulting ontological schemas of NELL-ZS and Wiki-ZS.

% \subsubsection{Schema Overview}
\subsection{Data Overview and Storage}
% \todo{[Add a conclusion, providing some extensive discussions (with evidence) on illustrate how our work is better than existing benchmarks]}

To the best of our knowledge, this resource is the first to incorporate with rich ontology information for tackling the ZS-KGC problems with unseen relations.
In comparison with the external knowledge contained in text, ontological schema provides richer and more accurate semantics about KG relations.
With our resources, various ontology-driven ZS-KGC methods are expected to develop.

Each ontological schema is saved in two formats.
The first is the original ontology file ended with ``.owl''. It can be directly loaded and easily viewed by ontology editors such as Protege.
The second is an RDF triple file to save the schema graph which is transformed from the ontology according to W3C OWL to RDF graph mapping.
It is convenient for graph embedding methods e.g., GNNs and KGE algorithms to process.
Note other mappings from OWL ontology to RDF graph can be considered by the user.

\section{Benchmarking and Results}\label{application}

\subsection{Evaluating ZSL Model Performance}

%We first show the usage of our resources in evaluating and comparing different ZSL methods under different external knowledge settings.
In this section, we evaluate and compare the performance of different ZSL methods under different KG settings, using the aforementioned resources.
We first introduce the ZSL methods and evaluation settings, and then separately introduce the results on ZS-IMGC, ZS-RE and ZS-KGC.

\subsubsection{ZSL Methods and Evaluation Settings}
As we introduce earlier, a kind of widely investigated ZSL methods are mapping-based.
They usually learn a mapping between the class embedding space and the sample space via data of seen classes, and generalize the learned mapping to unseen classes for prediction which is implemented by searching for matched classes for a testing sample according to some distance metrics.
According to the space where the searching is conducted, these methods can be further divided into three categories: semantic-space based which maps the sample feature to the space of class embedding, sample-space based which maps the class embedding to the space of sample feature, and common-space based which maps the sample feature and the class embedding to a common latent space.
%Semantic-space based methods learn a mapping function from the sample feature to the class embedding
%, and in testing, it searches for the nearest class in the class embedding space as the label of sample.
%Sample Common-space based methods embed sample features and class embedding into a common latent space.
In addition to these mapping-based methods, another popular branch of methods are generation-based. They formalize ZSL as a missing data problem and learn to synthesize samples (features) for unseen classes conditioned on their class embeddings to augment data.

In this work, we adopt two representative methods as our evaluation approaches.
One is a classic semantic-space based algorithm named DeViSE \cite{frome2013devise}, which is widely used in various ZSL studies.
The other method is a state-of-the-art generation-based method named OntoZSL \cite{geng2021ontozsl} which leverages Generative Adversarial Network (GAN) \cite{goodfellow2014generative} to generate data and is originally developed to utilize ontologies as external knowledge.
Another reason for selecting them is that they are compatible to different external knowledge that have been embedded, through which we are able to systematically compare different knowledge settings.
We benchmarked both methods for all the three ZSL tasks.
%applications we investigated in this paper.

It is worth noting that we considered the performance in two ZSL settings: one is the standard ZSL which tests samples of only unseen classes, the other is the generalized ZSL (GZSL) which tests samples of both seen and unseen classes and is more challenging.
%because  samples appear and the candidate label space involves both seen and unseen classes.
We evaluated under both settings for the ZS-IMGC and ZS-RE tasks, while for the ZS-KGC task, we only considered the standard ZSL setting since in our task of predicting a triple's object entity, the subject entity and the unseen relation are given (see more details in Section~\ref{experi_zskgc}). 
%considering that the triple completion with unseen relations will not be confused by the seen relations.
%More detailed explanations are provided in Section~\ref{experi_zskgc}. 

\subsubsection{ZS-IMGC}
With the KGs of the ZS-IMGC benchmarks, we made the following four external knowledge settings which have different semantics:
\begin{itemize}[leftmargin=0.4cm]
    \item \textbf{Basic KG}: Hierarchy and attribute triples which cover the semantics of class hierarchy, class attributes and attribute hierarchy.
    \item \textbf{Basic KG+literals}: Basic KG plus textual information.
    \item \textbf{Basic KG+CN}: Basic KG plus ConceptNet subgraph.
    \item \textbf{Basic KG+logics}: Basic KG plus disjointness axioms.
\end{itemize}
To apply these external knowledge in ZSL, we take advantage of some semantic embedding techniques to encode them and generate a vector representation for each class.
Specifically, we adopt mature and widely-used TransE \cite{bordes2013translating} to encode the graph structural knowledge contained in \textbf{Basic KG}, \textbf{Basic KG+CN} and \textbf{Basic KG+logics}.
For \textbf{Basic KG+literals}, we adopt a text-aware graph embedding method used in \cite{geng2021ontozsl} to simultaneously embed the textual and graph structural knowledge.

Besides the above KG-based external knowledge settings, we also made the following simple but widely used external knowledge settings. Relevant external knowledge can be extracted from the original benchmarks, and have also been included in our new resources:
\begin{itemize}[leftmargin=0.4cm]
    \item \textbf{Class Attributes (att)}: $85$, $85$ and $40$ dimensional binary-valued attribute vectors (multi-hot vectors) for classes of AwA, ImNet-A and ImNet-O, respectively.
    \item \textbf{Class Word Embeddings (w2v)}: Two kinds of word vectors for each class via its name. One is by \cite{changpinyo2016synthesized} with a dimension of $500$; the other is by a Glove model with a dimension of $300$.
    \item \textbf{Class Hierarchy (hie)}: A $100$ dimensional vector for each class, encoded by a graph auto-encoder \cite{kipf2016variational} over the class hierarchy.
\end{itemize}

\begin{table*}
\scriptsize
\caption{\small \textit{Acc}uracy (\%) of DeViSE on AwA, ImNet-A and ImNet-O. The best result on each metric is marked with underline.
}\label{results_imgc_devise}
\begin{tabular}{c|c|ccc|c|ccc|c|ccc}
% \hline
\toprule[0.5pt]
\multirow{2}{*}{\begin{tabular}[c]{@{}c@{}}External \\ Knowledge\end{tabular}}
&
\multicolumn{4}{c|}{AwA} 
&  \multicolumn{4}{c|}{ImNet-A}
&\multicolumn{4}{c}{ImNet-O}\\
% \cline{2-13}
& $acc$  & $acc_s$ & $acc_u$ & $H$ 
& $acc$  & $acc_s$ & $acc_u$ & $H$ 
& $acc$  & $acc_s$ & $acc_u$ & $H$ 
\\
% \hline
\midrule[0.3pt]
w2v (500)

& 24.22 & 78.42 & 1.05 & 2.08

& 13.52 & 59.71 & 0.63 & 1.25

& 14.21 & 66.40 & 3.93 & 7.43
\\
w2v (300)

& 8.42 & 86.32 & 0.00 & 0.00 

& 26.95 & \underline{84.36} & 0.16 & 0.32 

& 20.49 & \underline{93.60} & 0.00 & 0.00
\\
att
& 38.48 & 81.86 & 3.59 & 6.88

& \underline{35.72} & 61.00 & 12.60 & 20.89

& \underline{31.75} & 47.80 & 17.24 & 25.34
\\
hie
& 43.50 & 65.25 & 5.60 & 10.32

& 30.94 & 62.07 & 1.67 & 3.25

& 29.25 & 54.60 & 10.85 & 18.10
\\
% \hline
\midrule[0.3pt]
Basic KG  
& 43.24 & 86.44 & 6.40 & 11.91

& 34.38 & 25.50 & 28.13 & \underline{26.75}

& 24.77 & 34.20 & \underline{22.49} & 27.14
\\
Basic KG + literals
& \underline{46.12} & 84.42 & \underline{8.76} & \underline{15.88}

& 33.62 & 23.36 & \underline{29.33} & 26.01

& 26.13 & 38.60 & 21.67 & \underline{27.75}
\\
Basic KG + CN
& 45.56 & \underline{88.85} & 0.38 & 0.76

& 35.11 & 67.71 & 7.46 & 13.43

& 26.72 & 70.40 & 7.23 & 13.11 
\\
Basic KG + logics
& 37.54 & 80.69 & 1.09 & 2.15 
& -- & -- & -- & --

& -- & -- & -- & --
\\
% \hline
\bottomrule[0.5pt]
\end{tabular}
\end{table*}

\begin{table*}
\scriptsize
\caption{\small \textit{Acc}uracy (\%) of OntoZSL on AwA, ImNet-A and ImNet-O. The best result on each metric is also underlined.
}\label{results_imgc_ontozsl}
\begin{tabular}{c|c|ccc|c|ccc|c|ccc}
% \hline
\toprule[0.5pt]
\multirow{2}{*}{\begin{tabular}[c]{@{}c@{}}External \\ Knowledge\end{tabular}}
&
\multicolumn{4}{c|}{AwA} 
&  \multicolumn{4}{c|}{ImNet-A}
&\multicolumn{4}{c}{ImNet-O}\\
% \cline{2-13}
& $acc$  & $acc_s$ & $acc_u$ & $H$ 
& $acc$  & $acc_s$ & $acc_u$ & $H$ 
& $acc$  & $acc_s$ & $acc_u$ & $H$ 
\\
% \hline
\midrule[0.3pt]
w2v(500)
& 45.39 & 57.83 & 34.53 & 43.24

& 20.94 & 34.50 & 15.62 & 21.50

& 20.00 & 41.20 & 14.33 & 21.27
\\
w2v(300)
& 20.80 & 22.67 & 12.88 & 16.43

& 27.76 & 40.50 & 20.40 & 27.13

& 24.73 & 37.20 & 17.52 & 23.83
\\
att
& 58.47 & 59.90 & 44.24 & 50.89

& 37.87 & 33.50 & 27.62 & 30.28 

& 32.98 & 42.00 & 20.67 & \underline{27.71}
\\
hie
& 38.89 & 51.08 & 31.38 & 38.88

& 33.32 &  40.93 & 23.06 & 29.50 

& \underline{33.17} & 36.80 & \underline{21.13} & 26.85 
\\
% \hline
\midrule[0.3pt]
Basic KG  
& \underline{62.65} & 59.59 & \underline{50.58} & \underline{54.71}

& 38.21 & \underline{45.71} & 23.21 & 30.79

& 32.14 & 44.60 & 18.74 & 26.39
\\
Basic KG + literals

& 59.21 & 62.39 & 45.55 & 52.66
& \underline{38.58} & 35.64 & \underline{27.64} & 
\underline{31.13} 

& 32.57 & \underline{44.80} & 19.35 & 27.03
\\
Basic KG + CN
& 54.61 & 63.31 & 39.19 & 48.41

& 35.24 & 39.86 & 24.97 & 30.71

& 29.39 & 42.20 & 19.64 & 26.80
\\
Basic KG + logics
& 54.65 & \underline{65.37} & 40.76 & 50.21
& -- & -- & -- & --

& -- & -- & -- & --

\\
% \hline
\bottomrule[0.5pt]
\end{tabular}
\end{table*}

\begin{table*}
\scriptsize
\caption{\small \textit{Acc}uracy (\%) of GCNZ on AwA, ImNet-A and ImNet-O. The better result on each metric is marked with underline.
}\label{results_imgc_gcnz}
\begin{tabular}{c|c|ccc|c|ccc|c|ccc}
% \hline
\toprule[0.5pt]
\multirow{2}{*}{\begin{tabular}[c]{@{}c@{}}External \\ Knowledge\end{tabular}}
&
\multicolumn{4}{c|}{AwA} 
&  \multicolumn{4}{c|}{ImNet-A}
&\multicolumn{4}{c}{ImNet-O}\\
% \cline{2-13}
& $acc$  & $acc_s$ & $acc_u$ & $H$ 
& $acc$  & $acc_s$ & $acc_u$ & $H$ 
& $acc$  & $acc_s$ & $acc_u$ & $H$ 
\\
% \hline
\midrule[0.3pt]
Class Hierarchy

& 37.44 & \underline{81.45} & 7.86 & 14.34

& 33.95 & \underline{48.71} & 18.37 & 26.68

& 32.24 & \underline{49.00} & 18.04 & 26.37
\\
Basic KG
& \underline{62.98} & 75.59 & \underline{20.28} & \underline{31.98}

& \underline{36.64} & 45.57 & \underline{23.92} & \underline{31.38}

& \underline{33.98} & 43.80 & \underline{21.61} & \underline{28.94}
\\
% \hline
\bottomrule[0.5pt]
\end{tabular}
\end{table*}

We evaluate these external knowledge settings and two ZSL methods by macro accuracy following the standard in the ZS-IMGC community.
Macro accuracy is calculated in the following way: an accuracy, the ratio of correct predictions over all the testing samples, is first independently computed for each class, and then the accuracies of all tested classes are averaged.
In the stardard ZSL setting, we compute the macro accuracy over all the unseen classes, denoted as $acc$.
In the generalized ZSL setting, two macro accuracies are calculated over the seen classes and the unseen classes, respectively, denoted as $acc_s$ and $acc_u$, and a harmonic mean $H=(2 \times acc_s \times acc_u)/(acc_s+acc_u)$ which balances the performance of predicting seen classes and unseen classes is calculated as the overall metric.

The results of DeViSE and OntoZSL on three datasets are shown in Table~\ref{results_imgc_devise} and Table~\ref{results_imgc_ontozsl}, respectively.
From both tables, we can see that the KG-based external knowledge settings all achieve better performance than those currently widely used none-KG-based settings --- \textbf{att}, \textbf{w2v} and \textbf{hie} on AwA.
Although the KG-based settings do not always achieve the best performance w.r.t. all the metrics on ImNet-A and ImNet-O, the results are still comparable.
All these illustrate the potential of KG-based external knowledge in ZS-IMGC.
Besides, with TransE which was originally developed for KGs with relational facts alone, and a simple pipeline that stacks semantic embedding and a ZSL method, although more semantics are introduced in \textbf{Basic KG+CN} and \textbf{Basic KG+logics}, their results are not better than \textbf{Basic KG} on most metrics.
This motivates the community to \textit{i)} develop more effective ZS-IMGC techniques to utilize all these promising semantics for better performance, or \textit{ii)} adopt more flexible strategies to take advantage of these semantics such as retrieving more refined knowledge for specific prediction tasks or datasets, and our resources provide a chance for such investigations.

We also observe the performance differences when applying the setting of w2v to different datasets, i.e., \textbf{w2v(500)} performs better than \textbf{w2v(300)} on AwA, whereas on ImNet-A/O, the situation is inverse --- \textbf{w2v(300)} performs better on most metrics.
For example, when experimenting with OntoZSL, on AwA, \textbf{w2v(500)} achieves $45.39\%$ in $acc$ and $43.24\%$ in $H$, with improvements of $24.59\%$ in $acc$ and $26.81\%$ in $H$ over \textbf{w2v(300)}.
While for ImNet-A, on the metrics of $acc$ and $H$, \textbf{w2v(300)} inversely achieve respective $6.82\%$ and $5.63\%$ performance gains against \textbf{w2v(500)}.
For ImNet-O, the improvements are $4.73\%$ on $acc$ and $2.56\%$ on $H$.
These two kinds of word embeddings are both pre-trained on Wikipedia corpus, but the difference is that \textbf{w2v(500)} by \cite{changpinyo2016synthesized} directly learns a word vector for each class, no matter the class name contains a single word or multiple words, while \textbf{w2v(300)} generates word vectors for multiple-word classes by averaging the word vectors of terms in names.
The averaged class word embeddings may be more beneficial for the fine-grained classes of ImNet-A/O as the words in the names of the classes in a family are highly overlapped.
Instead, the classes in AwA are coarse-grained, the overlap of words in their names is relatively less.  
Correspondingly, in encoding the text in the \textbf{Basic KG+literals} with 300-dimensional word vectors from Glove (i.e., \textbf{w2v(300)}), \textbf{Basic KG+literals} gains an improved performance w.r.t most metrics against \textbf{Basic KG} on ImNet-A/O whereas even performs worse on AwA when experimenting with OntoZSL.
See the fifth and sixth rows of Table~\ref{results_imgc_devise} and Table~\ref{results_imgc_ontozsl} for more detailed comparisons.
According to these observations, we can conclude that \textbf{w2v(300)} may be more appropriate for tasks with fine-grained classes, while \textbf{w2v(500)} may be better for tasks with coarse-grained classes.
Moreover, it also inspires us to take the properties of the task and data into account when designing semantic embedding techniques for the external knowledge.

% \todo{Comparison of different ZSL methods.}

%In addition, 
Apart from these two representative ZSL methods, we also evaluate a famous KG-based ZSL method named GCNZ \cite{wang2018zero}, which uses a Graph Convolutional Network (GCN) to propagate features between class nodes and outputs a classifier for each unseen class, under two semantic settings that GCNZ can support, i.e., \textbf{Basic KG} and \textbf{Class Hierarchy}.
GCNZ is a typical method of the propagation-based paradigm which has attracted wide attention for KG-based ZS-IMGC studies \cite{chen2020zero,gao2019know,wei2019residual,nayak2020zero,geng2020explainable}.
%Its manner is widely adopted by many ZS-IMGC studies which seek to infuse KG into the models \cite{chen2020zero,gao2019know,wei2019residual,nayak2020zero,geng2020explainable}.
%Here, we mainly make a comparison between our \textbf{Basic KG} and the semantic of \textbf{Class Hierarchy} which is originally used in GCNZ.
The results are shown in Table~\ref{results_imgc_gcnz}.
It can be seen that along with more semantics introduced in \textbf{Basic KG} than in \textbf{Class Hierarchy}, a higher performance is obtained over all three datasets, including an improvement on predicting unseen classes in the standard ZSL and a better balance between $acc_s$ and $acc_u$ in the generalized ZSL.
We also find a larger performance gap between \textbf{Basic KG} and \textbf{Class Hierarchy} on AwA than on ImNet-A/O.
It may be because the distances between AwA classes in the class hierarchy graph are larger since the nature of coarse granularity of AwA, and this may penalize the feature propagation in the graph convolutional layer.
Contrastingly,
when more semantics such as class attributes are introduced in \textbf{Basic KG}, the connections between AwA classes are greatly enriched, resulting in a significant performance improvement.
It is worth noting that though we ignore the relational edges in the \textbf{Basic KG} when GCNZ is applied (since GCN can only support single relation graph), we still obtain promising results.
In the future, more GCN variants that can support multi-relation graphs such as R-GCN \cite{schlichtkrull2018modeling} and CompGCN \cite{vashishth2019composition} can be investigated to make full use of the semantics of the KG.

\begin{table}
\scriptsize
  \caption{\small \textit{Acc}uracy (\%) of DeViSE and OntoZSL on ZeroRel. The best result on each metric is underlined.}
  \label{results_on_ZeroRel}
% \begin{tabular}{c|c|ccc|c|ccc}
\begin{tabular}{p{1.0cm}<{\centering}|p{0.5cm}<{\centering}|p{0.4cm}<{\centering}p{0.4cm}<{\centering}p{0.5cm}<{\centering}|p{0.5cm}<{\centering}|p{0.4cm}<{\centering}p{0.4cm}<{\centering}p{0.5cm}<{\centering}}
% \hline
\toprule[0.5pt]
% \multirow{2}{*}{\begin{tabular}[c]{@{}c@{}}CE \\ Variants\end{tabular}} 
\multirow{2}{*}{Semantics
% \begin{tabular}[c]{@{}c@{}}External \\ Knowledge\end{tabular}
}
& \multicolumn{4}{c|}{DeViSE} &  \multicolumn{4}{c}{OntoZSL} \\
% \cline{2-13}
& $acc$  & $acc_s$ & $acc_u$ & $H$
& $acc$  & $acc_s$ & $acc_u$ & $H$ \\
% \hline
\midrule[0.3pt]
w2v 
% & 13.52 & 59.71 & 0.63 & 1.25   & 20.87 & 40.14 & 13.90 & 20.65 
&21.40 &\underline {65.73} &0.04&0.08
&11.00&\underline {56.86}&0.20&0.40
\\
KG
% & 26.95 & {\bf 84.36} & 0.16 & 0.32   & 27.76 & 40.50 & 20.40 & 27.13 
&34.43&63.64&7.47&13.37
&35.81&53.06&9.21&15.69
\\
Rule
% & {\bf 35.39} & 77.07 & 4.21 & 7.98  
% & 37.87 & 40.71 & 24.00 & 30.20 
& \underline{35.74} & 61.96& \underline{10.20}& \underline{17.51}
&\underline{36.20} &52.20&\underline{12.05}&\underline{19.58}
\\
% \hline
\bottomrule[0.5pt]
\end{tabular}
\end{table}

\subsubsection{ZS-RE}
We evaluate one baseline external knowledge setting which is based on word embeddings, and other two external knowledge settings based on our resources introduced in Section \ref{zsre_semantics}:
%kinds of external knowledge of relation types developed in Section~\ref{zsre_semantics} and compare them with a baseline semantic setting from word embeddings:
\begin{itemize}[leftmargin=0.4cm]
    \item \textbf{w2v}: one 100-dimensional word vector for each relation by averaging the words in its name. The word embedding model is trained on the latest Wikipedia dump\footnote{\url{https://dumps.wikimedia.org/enwiki/latest/enwiki-latest-pages-articles.xml.bz2}} using the word2vec algorithm \cite{mikolov2013distributed} with a window size of 5.
    \item \textbf{KG}: one $100$-dimensional KG embedding for each relation. The KG entity and relation embeddings are trained by the OpenKE \cite{han2018openke} toolkit using TransE on the Wikidata dump constructed in Section~\ref{zsre_semantics}.
    %, which uses TransE to train $100$-dimensional entity and relation embeddings on the Wikidata dump.
    \item \textbf{Rule}: one 100-dimensional rule-guided relation embedding for each relation. The embedding method incorporated with rules are introduced bellow.
\end{itemize}

We leverage the pre-trained KG embeddings for initial relation embeddings, and then utilize the extracted rules to generate rule-guided relation embeddings.
For a length-1 rule $r_1(x, y) \Rightarrow r_2(x, y)$ and a length-2 rule
$r_3 (x, y) \wedge r_4 (y, z) \Rightarrow r_5 (x, z)$, we can get $\vec{r_1} = \vec{r_2}$ and $\vec{r_3} + \vec{r_4} = \vec{r_5}$ according to the TransE KGE algorithm which assumes $\vec{s}+\vec{r} \approx \vec{o}$ for a valid triple $(s, r, o)$.
Note $\vec{r_1}, \vec{r_2}, \vec{r_3}, \vec{r_4}, \vec{r_5}, \vec{s}, \vec{r}, \vec{o}$ all represent entity and relation embeddings. 
Thus, for a relation associated with such rules, its embedding can be re-calculated based on the embeddings of other relations in the rules.
Specifically, the embedding of relation $r$ associated with $K$ rules is calculated as follows:
\begin{equation}\nonumber
    E_{rl} (r) = \frac{\sum_{k=1}^{K} s_k * E_{TransE} (R_k^r) }{\sum_{k=1}^{K} s_k}
\end{equation}
where $R_k^r$ is the $k$-th rule of relation $r$, and $s_k$ corresponds to its confidence score.
$E_{TransE}(\cdot)$ represents the rule-based operation following the TransE's assumption.
For example, for an unseen relation $r_u$ with its three rules, $R_1$: $r_A \Rightarrow r_u$, 
$R_2$: $r_B \wedge r_C \Rightarrow r_u$, and 
$R_3$: $r_D \wedge r_u \Rightarrow r_E$, its rule-guided embedding is calculated as:  
$$
    E_{rl} (r_u) = \frac{s_1* \vec{r_A} + s_2*(\vec{r_B} + \vec{r_C})+s_3 * (\vec{r_E} - \vec{r_D})}{s_1+s_2+s_3}.
$$

% As for the word embeddings, we train 100-dimensional word vectors on the latest Wikipedia dump corpus\footnote{\url{https://dumps.wikimedia.org/enwiki/latest/enwiki-latest-pages-articles.xml.bz2}} using word2vec algorithm \cite{mikolov2013distributed} with a window size of 5.

Apart from the ZSL method of DeViSE which had been evaluated in \cite{li2020logic}, in this paper, we also extend to experiment with OntoZSL, i.e., learning to generate instance features (extracted by Piecewise Convolutional Neural Networks \cite{zeng2015distant}) for unseen relations conditioned on their semantic embeddings.
Both methods are assembled with the three external knowledge settings mentioned above. 
The results on our newly constructed dataset ZeroRel are shown in Table~\ref{results_on_ZeroRel}, where the classification accuracy is reported.
Similar to ZS-IMGC, we compute the accuracy independently for each relation type, and report the averaged accuracy on unseen relations in the standard ZSL (i.e., $acc$) and on seen and unseen relations in the generalized ZSL (i.e., $acc_s$ and $acc_u$, respectively) with their harmonic mean computed.

From Table~\ref{results_on_ZeroRel}, we can see that the settings of \textbf{KG} and \textbf{Rule} both achieve significant improvements over the \textbf{w2v} setting, no matter what ZSL methods are used.
Especially, the \textbf{Rule} setting leads to the best performance.
All of these results illustrate that the semantics in the external knowledge from KG and logic rules
are richer than the semantics in the word embeddings w.r.t. the task of ZS-RE, and the rule-guided embedding method we used is effective.
In the future, we plan to develop more techniques to utilize the semantics in KGs and the logic rules.
%rather than being limited to the pre-trained KG embeddings.
%
We also find that the generation-based method OntoZSL performs better than the mapping-based method DeViSE, especially with respect to $acc_u$ and $H$, under the knowledge settings of \textbf{KG} and \textbf{Rule} which contain richer semantics than \textbf{w2v}.
For example, under the setting of \textbf{Rule}, the $H$ score of OntoZSL is $11.8\%$ higher than that of DeViSE.

% \begin{itemize}[leftmargin=0.4cm]
%     \item \textbf{Word Embeddings of Relations (w2v)}:
%     \item
%     \item
% \end{itemize}

\begin{table*}[]
\scriptsize
\caption{\small Results (\textit{MRR} and \textit{hit@k} (\%)) of DeViSE and OntoZSL on NELL-ZS and Wiki-ZS. The best result on each metric is underlined.}
\label{kgc_results}
\begin{tabular}{c|cccc|cccc|cccc|cccc}
% \hline
\toprule[0.5pt]
\multirow{4}{*}{\begin{tabular}[c]{@{}c@{}}External\\ Knowledge\end{tabular}} & \multicolumn{8}{c}{\small \textbf{DeViSE}}  
& \multicolumn{8}{c}{\small \textbf{OntoZSL}}
\\
& \multicolumn{4}{c|}{\textbf{NELL-ZS}} 
& \multicolumn{4}{c|}{\textbf{Wiki-ZS}}      & \multicolumn{4}{c|}{\textbf{NELL-ZS}}
& \multicolumn{4}{c}{\textbf{Wiki-ZS}} 
\\
& \multirow{2}{*}{MRR} & \multicolumn{3}{c|}{hit@} & \multirow{2}{*}{MRR} & \multicolumn{3}{c|}{hit@} & \multirow{2}{*}{MRR} & \multicolumn{3}{c|}{hit@} & \multirow{2}{*}{MRR} & \multicolumn{3}{c}{hit@} \\\cline{3-5}\cline{7-9}\cline{11-13}\cline{15-17}
&   & 10      & 5      & 1     &                      & 10      & 5      & 1     &                      & 10      & 5      & 1     &                      & 10      & 5      & 1     \\
% \hline
\midrule[0.3pt]
Text              
& 0.221 & 34.6 & 29.0 & 15.5

& 0.183 & 26.7 & 21.7 & 13.5

& 0.215 & 34.5 & 28.3 & 14.5
& 0.185 & 27.3 & 22.3 & 13.5
\\
% \hline
\midrule[0.3pt]
\multicolumn{1}{l|}{\ RDFS-hie}
& \underline{0.229} & 35.1 & 29.3 & \underline{16.3}

& 0.179 & 25.4 & 20.9 & 13.5

& 0.225 & 34.8 & 28.9 & \underline{15.9}

& 0.175 & 25.4 & 20.4 & 13.1
\\
\multicolumn{1}{l|}{\  RDFS-cons}
& 0.221 & 34.5  & 28.7 & 15.3

& 0.183 & 26.4 & 21.7 & 13.6

& 0.220 & 34.3 & 28.0 & 15.4

& 0.177 & 25.7 & 21.2 & 13.0
\\
\multicolumn{1}{l|}{\  RDFS graph}
& 0.225 & \underline{35.3} & \underline{29.4} & 15.6

& 0.184 & 27.0 & 21.7 & 13.6

& 0.223 & 35.1 & 29.1 & 15.3

& 0.185 & 27.5 & 22.3 & 13.4 

\\
% \hline
\midrule[0.3pt]
RDFS+literals
& 0.223 & 35.0 & 29.0 & 15.3

& \underline{0.185} & \underline{27.1} & \underline{22.0} & \underline{13.6}

& \underline{0.227} & \underline{35.6} & \underline{29.4} & 15.6 

& \underline{0.188} & \underline{28.1} & \underline{22.6} & \underline{13.5}
\\
% \hline
\bottomrule[0.5pt]
\end{tabular}
\end{table*}

\subsubsection{ZS-KGC}\label{experi_zskgc}
For NELL-ZS and Wiki-ZS, we made one simply none KG external knowledge setting that has already been included in the original benchmark, and five KG-based settings using our new resources:
\begin{itemize}[leftmargin=0.4cm]
    \item \textbf{Text}: relations' textual descriptions which are originally proposed and used by Qin et al. \cite{qin2020generative};
    \item \textbf{RDFS graph}: complete relation semantics in RDFS;
    \item \textbf{RDFS-hie}: a part of RDFS graph, covering subproperty and subclass triples;
    \item \textbf{RDFS-cons}: a part of RDFS graph, covering relation domain and range constraints; 
    \item \textbf{RDFS+literals}: RDFS graph plus textual meta data of concepts and relations;
    \item \textbf{RDFS+OWL}: RDFS graph plus semantics in OWL. 
\end{itemize}

As in ZS-IMGC and ZS-RE, we embed the external knowledge and apply the resultant relation vectors into the ZS-KGC methods.
For \textbf{RDFS graph} and its two subgraphs (\textbf{RDFS-hie} and \textbf{RDFS-cons}), we adopt TransE for embedding, and for \textbf{RDFS+literals}, we also adopt the text-aware graph embedding method used in \cite{geng2021ontozsl}.
To embed the relation's textual descriptions, we follow \cite{qin2020generative} and perform a weighted summation of the vectors of words in descriptions, where an open word embedding set\footnote{\url{https://github.com/mmihaltz/word2vec-GoogleNews-vectors}} of dimension 300 is used.
We also use the same word vectors to initialize the representation of text in \textbf{RDFS+literals}.

We first compare these five external knowledge settings using methods of DeViSE and OntoZSL, where the sample features of KG relations are learned by their associated entity pairs. 
% and will be used for  both of which \todo{learn to align the sample features of relations (i.e., the representations of associated entity pairs) and the embeddings of relations' external knowledge.}
The KGC task here is to predict the object entity given a subject entity and a relation, we thus rank all the candidate entities according to their likelihood to be the object.
Two commonly used metrics are adopted: mean reciprocal ranking (\textit{MRR}) which computes the average of the reciprocal predicted ranks of all the ground truths (right objects),
and $hit@k$ which represents the ratio of testing samples whose ground truths are ranked in the top-$k$ positions ($k$ is set to 1, 5, 10) \cite{wang2017knowledge}.
As the candidate space only involves entities, the prediction with unseen relations is independent of the prediction with seen relations, and in fact, the latter is the traditional KGC task. Therefore, we only consider the standard ZSL testing setting in ZS-KGC.
The results are shown in Table~\ref{kgc_results}.

From Table~\ref{kgc_results}, we find that in comparison with \textbf{Text}, \textbf{RDFS graph} and \textbf{RDFS+literals} always lead to better performance, \textbf{RDFS-hie} and \textbf{RDFS-cons} that contain incomplete RDFS semantics also achieve comparable results, and even perform better on some metrics.
For example, when applying the OntoZSL method on NELL-ZS, the \textit{MRR} values of \textbf{Text}, \textbf{RDFS-hie}, \textbf{RDFS-cons}, \textbf{RDFS graph} and \textbf{RDFS+literals} are $0.215$, $0.225$, $0.220$, $0.223$ and $0.227$, respectively. 
These results demonstrate the superiority of our proposed RDFS-based relation semantics.
Besides, we also find that when changing the external knowledge from \textbf{Text} to \textbf{RDFS graph}, a higher improvement is achieved on NELL-ZS than on Wiki-ZS.
For example, when OntoZSL is applied, the value of $hit@10$ is improved by 0.6\% on NELL-ZS, whereas is only improved by 0.2\% on Wiki-ZS. And on Wiki-ZS, the results w.r.t other metrics are roughly the same. See the first and fourth row of Table~\ref{kgc_results} for more comparisons.
These results are mainly due to that 44 out of 537 Wikidata relations have missing semantics in RDFS.
When complementing the RDFS graph with relations' textual information (i.e., \textbf{RDFS+literals}), the performance on Wiki-ZS is further improved.
% \todo{higher performance} is achieved.

In Table~\ref{kgc_results}, it can also be seen that \textbf{RDFS-hie} performs better than \textbf{RDFS-cons} on NELL-ZS, while \textbf{RDFS-cons} inversely achieves better performance on Wiki-ZS.
For example, when experimenting with DeViSE, the \textit{MRR} value of \textbf{RDFS-hie} on NELL-ZS is $0.229$ and $3.6\%$ relative higher than that of \textbf{RDFS-cons}, while on Wiki-ZS, the \textit{MRR} is decreased by $0.004$ when shifting the setting from \textbf{RDFS-cons} to \textbf{RDFS-hie}.
This is probably because for all relations in the NELL-ZS, around 58\% of them are hierarchically related, while only nearly 30\% have identical domain and range constraints; while in Wiki-ZS, the constraint information are richer than the hierarchy information, i.e., 160 relations are hierarchically related, and although only 161 relations have the same domain and range constraints, 
over 70\% of them have more than one identical domains or ranges.
% \todo{and although only 161 relations have at least one pair of 
% the same domain and range constraints, nearly 28\% of them have at least two pairs of identical domain and range constraints.}

The external knowledge defined by OWL is quite promising for augmenting ZSL, but there are current no systemic or robust methods. 
We try to validate the effectiveness of OWL semantics by testing an ensemble method, which combines symbolic reasoning and embedding-based prediction, under the setting of \textbf{RDFS+OWL}.
Briefly, for a testing tuple (subject and relation), if the object can be inferred through logical expressions (cf. Section~\ref{OWL_Axioms}), we adopt the inferred object; otherwise, we use the predicted object ranking by ZSL models such as OntoZSL.
Even with such a naive ensemble solution, we got some encouraging results. 
On Wiki-ZS, $5$ unseen relations have inverse relations which are among the seen relation set, and $hit@1$ increases from $28.6\%$ to $55.3\%$ when logical inference with the inverse semantics is used.
On NELL-ZS, $4$ unseen relations are composed by $10$ seen relations, and the logical inference with composition leads to $6.3\%$ increment on $hit@1$.
These results demonstrate the superiority of the OWL-based relation semantics, although the method of utilizing OWL axioms presented here is quite preliminary.
With our resources, more robust methods can be investigated to better utilize such logical expressions and significantly augment ZSL performance.

Besides, the OWL semantics can also be used to validate the correctness of prediction results or improve the prediction efficiency.
For example, irreflexivity  constrains that the predicted object entity can not be the same with the given subject entity; and both irreflexivity and asymmetry can be used to reduce the searching space of candidate entities during prediction.
In the future, more available and effective methods are expected to take advantage of these semantics for promoting ZSL.

\subsection{Evaluating ZSL Model Explanation}
Argumentative machine learning explanation by additional domain or background knowledge has been widely investigated \cite{vcyras2021argumentative,chen2018knowledge,confalonieri2021using}.
Our ZSL resources with rich external knowledge can also be used to investigate explainable ZSL methods and evaluate ZSL methods' interpretation and prediction justification.
For demonstration, we use different external knowledge settings that can be made by our resources to evaluate a knowledge augmented ZSL explanation method named X-ZSL which explains the transferability of sample features in zero-shot image classification in a human understandable manner \cite{geng2020explainable}.
Briefly, X-ZSL first uses an Attentive Graph Neural Network to automatically learn seen classes that are important to the feature learning of an unseen class, then explains the feature transfer between them by extracting class knowledge from KGs, and finally uses some templates to generate human understandable natural language explanations.

\begin{figure*}
\centering
\includegraphics[width=0.98\textwidth]{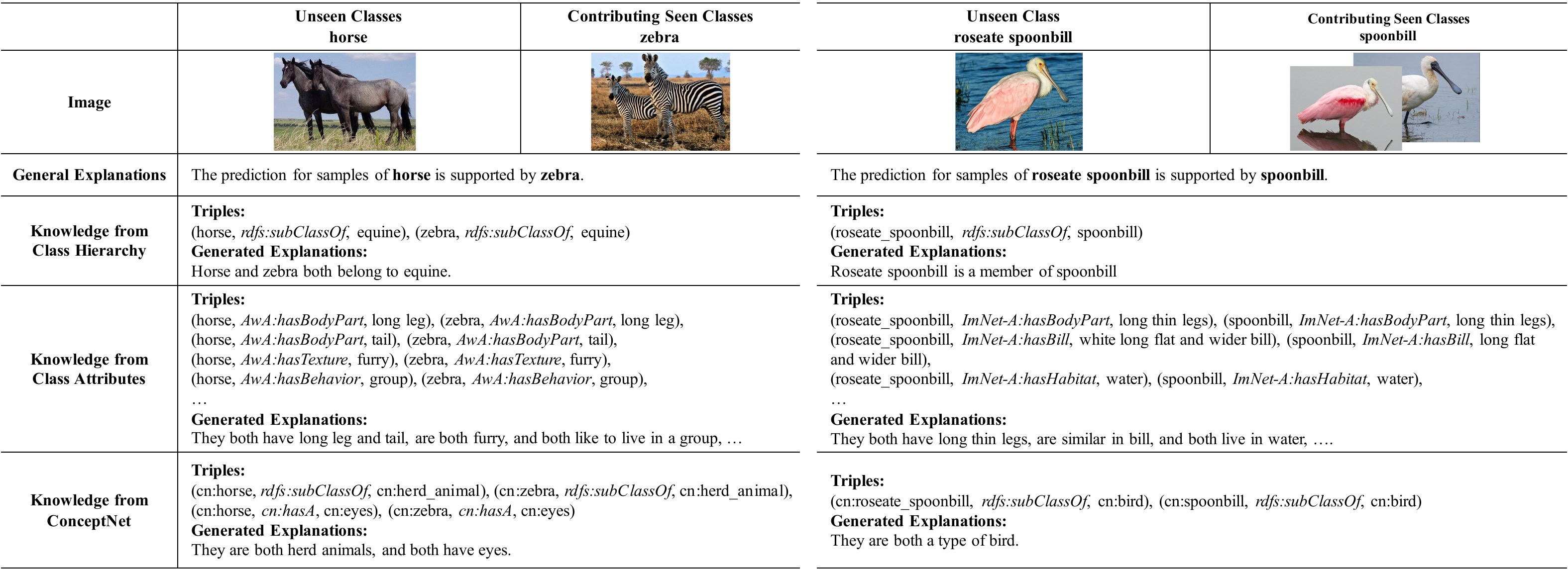}
%\vspace{-0.2cm}
\caption{\small Examples for explaining why features of seen classes \textit{zebra} and \textit{spoonbill} are transferred to unseen classes \textit{horse} and \textit{roseate spoonbill}, respectively.
Note we have replaced human unreadable entity ids by entity names.}\label{explanation_example}
\end{figure*}

Fig.~\ref{explanation_example} presents and compares X-ZSL's explanations using different external knowledge.
We give two examples, each of which includes one unseen class and one seen class that contributes to this unseen class, from AwA and ImNet-A --- two benchmarks for ZS-IMGC.
%on some animal classes from AwA and ImNet-A in the ZS-IMGC case, with different knowledge extracted from our KG resources.
As we can see, for the feature transferability between seen and unseen classes, the knowledge from class hierarchy provides overall explanations, from the perspective of their relatedness in biology; the knowledge from class attributes provides detailed explanations, from the perspective of their relatedness in characteristics especially in visual characteristics; while the relational facts from ConceptNet provide an important supplement.
In summary, different semantics in our resources all can have positive contributions to explain a ZSL model or to justify a ZSL prediction, and thus more explanation methods with different manners can be investigated and compared with using our resources.

\section{Discussion and Outlook} \label{sec:discussion}

\subsection{ZSL Methods}

%In evaluating the impact of different external knowledge settings on ZSL prediction performance,
For all the knowledge settings across all the tasks,
we compared two general ZSL methods --- DeViSE and OntoZSL. 
%across different ZSL tasks.
According to the results, we find that the generation-based method OntoZSL always has superior performance than the mapping-based method DeViSE no matter in the standard ZSL setting or in the generalized ZSL setting, especially in ZS-IMGC and ZS-RE.
For example, on AwA in ZS-IMGC, DeViSE achieves average $35.89\%$  $acc$ and average $6.25\%$ $H$ across all the knowledge settings, whereas the average $acc$ and $H$ values of OntoZSL are $49.33\%$ and $44.43\%$, respectively.
This may due to the hubness problem \cite{shigeto2015ridge} that exists in the label searching of DeViSE.
That is, since DeViSE maps a number of sample features to a point in the class embedding space for a certain class, during prediction, this manner will increase the probability of irrelevant points (hubs) being the nearest neighbors (i.e., the matched classes or relations).
In contrast, OntoZSL takes a different strategy which generates a number of unseen samples (features) so that classifiers can be trained to classify the unseen testing samples.
However, in the ZS-KGC task, DeViSE performs better than OntoZSL on some metrics.
This is because we conduct an inverse mapping in the ZS-KGC case, i.e., mapping one semantic representation to a number of sample features for a relation, as the relation semantics have higher dimensions.
This can suppress the hubness problem to some extent.

In particular, we also find that in the generalized ZSL, the performance gap between DeViSE and OntoZSL is larger than in the standard ZSL. 
This is because the mapping-based methods are trained only by the samples of seen classes, and thus have a strong bias towards seen classes during the prediction of generalized ZSL, while for the generation-based methods, they convert the ZSL problem to a standard supervised learning problem and thus the bias toward unseen classes in prediction is avoided. 

Although OntoZSL usually achieves better performance, there is an issue that can not be overlooked -- OntoZSL contains more parameters compared with DeViSE, and thus it takes much more effort to search for appropriate parameters and hyper parameters such as the dimension of random noise vectors and the number of synthesized samples.
% \todo{Therefore, in practice, we need to select a suitable method according to the scenario we encounter.
% DeViSE can be used as a preliminary model to feel out the zero-shot scenario, while OntoZSL can be used to achieve higher prediction results with fine parameter engineering.}

Apart from these two ZSL methods that are applicable to all the tasks and all the external knowledge settings, we also evaluated a ZSL method of the propagation-based paradigm named GCNZ in ZS-IMGC to exploit the single relation graph of the KG.
%structured external knowledge and compare their performances.
In addition to GCN, other multi-relation graph neural networks such as R-GCN \cite{schlichtkrull2018modeling} and CompGCN \cite{vashishth2019composition} can also been applied and evaluated.
Actually, methods of the propagation-based paradigm have been rarely applied and compared in tasks beyond ZS-IMGC, such as ZS-RE and ZS-KGC.
Our resources make this possible, and we regard this as an important future work for the ZSL community.
%tasks that also contains semantics in the form of graph, we leave this as an important future work.
There are also some recent works leveraging pre-trained language models (PLMs) such as BERT \cite{devlin2019bert} to tackle the zero-shot problems in ZS-RE \cite{chen2021zs} and ZS-KGC \cite{yao2019kg,wang2021structure}.
Coupled with semantics from large amount of free text data, these models can easily generalize to unseen elements with the textual external knowledge.
Recently, there are also a variety of studies trying to integrate structured knowledge such as KGs into the current language models to leverage both structured and unstructured semantics, this provides us an opportunity to combine the ZSL methods of the PLM-based paradigm with our constructed KG resources.
Moreover, impressive multi-modal pre-training techniques can also be experimented for the ZS-IMGC tasks.

% In addition to develop more robust ZSL methods, in the future, we also plan to make a further systemic evaluation towards many ZSL methods, especially those utilizing KGs, using our constructed external knowledge and benchmark resources.

Reviewing all the experiment results, it can also be observed that although our proposed external knowledge settings achieve great balance between $acc_s$ and $acc_u$ metrics in the generalized ZSL, they do not always work well on $acc_s$.
This motivates us to explore more robust ZSL methods to maintain the high prediction accuracy on seen labels in the generalized ZSL setting.
% predict unseen testing samples correctly as well as maintain reasonable accuracy on seen labels.

% \todo{And, the benchmarks are also expected to be split according to other different criteria so that challenging these ZSL methods from multiple aspects.}

\subsection{External Knowledge}

In Section~\ref{application}, we evaluated different external knowledge settings including none KG-based ones such as \textbf{w2v} in ZS-IMGC and ZS-RE and KG-based ones with varying semantics.
%developed by ours with various semantics.
We find KG-based settings always achieve better prediction performance, especially in ZS-RE and ZS-KGC where KG-based external knowledge have been rarely studied.
%the results of the KG-based settings are especially impressive.
Moreover, KG settings with richer semantics usually have better performance.
For example, \textbf{Basic KG+literals} has better performance than \textbf{Basic KG} in ZS-IMGC, and \textbf{Rule} has better performance than \textbf{KG} in ZS-RE.
%and \textbf{RDFS+literals} in ZS-KGC lead to better performance than 
%such as \textbf{Basic KG+literals} in ZS-IMGC, \textbf{Rule} in ZS-RE, and \textbf{RDFS+literals} in ZS-KGC.
All these validate our motivations of investigating various external knowledge via KGs.

Although promising performance has been achieved, there are still some open problems w.r.t. utilizing the external knowledge.
%some derivative questions that deserves to be studied.
First, more advanced semantic embedding techniques are required to jointly embed different kinds of KG semantics.
%utilize the semantics contained in the constructed knowledge.
We adopted some simple semantic embedding techniques, such as TransE, and developed some new techniques for text-aware and logic-aware KG embedding.
%method and a logic-guided embedding method, to infuse them into the ZSL models.
Although they are quite effective when combined with OntoZSL and DeViSE, more advanced methods could be developed for better performance and for addressing some knowledge settings that the current embedding methods cannot address, such as \textbf{Basic KG+CN} and \textbf{Basic KG+logics} of ZS-IMGC and \textbf{OWL} of ZS-KGC.
%Although the effectiveness of these techniques have been validated in most of settings, they are still limited to represent KGs well, especially those with more semantics such as \textbf{Basic KG+CN} and \textbf{Basic KG+logics} for  ZS-IMGC, or even they can not represent some KGs with complex semantics, such as \textbf{OWL} for ZS-KGC.
We expect that some more complicated semantic embedding methods, such as multi-relation graph embedding \cite{schlichtkrull2018modeling,vashishth2019composition}, ontology embedding \cite{chen2020owl2vec} and multi-modal KG embedding \cite{gesese2019survey}, can be evaluated in combination with different ZSL methods.
%are developed to better and fully utilize all kinds of external knowledge.
% Besides, the impact of the dimensionality of external knowledge embeddings also should be considered and evaluated.

%The second is how to adaptively take advantage of these knowledge.
Second, more adaptive solutions are required for some specific KG-based knowledge settings, besides the current pipeline of first embedding the KG and then applying the class embedding in an existing ZSL method.
This is motivated by the observation that sometimes some specific knowledge settings have different performance on different datasets; for example,  \textbf{RDFS-hie} has better performance than \textbf{RDFS-cons} on NELL-ZS, but has worse performance than \textbf{RDFS-cons} on Wiki-ZS.
%and \textbf{w2v(300)} and \textbf{w2v(500)} for AwA and ImNet-A/O.
It seems that different datasets have different knowledge preferences, and it is necessary to take the properties of datasets into account when utilizing the same external knowledge.
% \todo{For example, we can adopt an embedding fusion strategy to embed different KG components using different but more suitable embedding methods, such as embedding the hierarchical knowledge using representative hierarchical graph embedding method e.g., HAKE \cite{zhang2020learning}, and then attentively fuse them according to the data properties.}
One promising solution is first automatically retrieving dataset-relevant knowledge from the KG and then utilizing these knowledge in ZSL.
Some learnable knowledge retrieval solutions could be considered in the future \cite{izacard2020distilling}.
%Other alternatives include a knowledge retrieval strategy that can manually or automatically retrieve knowledge relevant to the tasks or datasets from the resources.

Logical expressions are all considered in three ZSL tasks, i.e., disjointness axioms in ZS-IMGC, logic rules in ZS-RE and OWL axioms in ZS-KGC.
Their incorporation in different ZSL datasets and different ZSL methods also needs more adpative methods.
Compared with ZS-IMGC, ZS-RE and ZS-KGC seem to benefit more from these logic expressions.
This may be because the latter two are both symbolic inference tasks, i.e., inferring the semantic relations between two entity mentions and inferring the object entities given the subjects and relations.
%the symbolic logics may be more helpful.
For the ZS-IMGC task, there exists a gap between the symbolic class knowledge and class instances (i.e., images), and it is more challenging to utilize the logic expressions.
%Of course, it is obvious that the method of utilizing the logic expressions in ZS-IMGC is not perfect now. 
%Therefore, in the future, more effective methods are expected to adaptively utilize logics in different ZSL tasks and different ZSL datasets.
%to bridge the gap between symbolic knowledge and neural feature learning.
%Besides, the dimensionality of the embeddings of different external knowledge studied in the same ZSL task differs greatly, we leave comparing them systematically as a future work.

Third, high quality and large scale KG-based resources covering more tasks and more knowledge settings are required.
We plan to continuously extend our resources in the following aspects:
%to support the following needs in order:
\textit{i)} more external knowledge will be added for the existing ZSL tasks, such as more logical expressions for ImNet-A/O and AwA;
%\todo{such as enriching the ontological schemas for NELL-ZS and Wiki-ZS with more complex logical relationships;}
\textit{ii)} resources of more tasks, such as zero-shot visual question answering that we are investigating \cite{chen2021zero}, will be added;
\textit{iii)} more ZSL settings, such as ZSL with incremental unseen classes \cite{wei2021incremental}, will be considered;
and \textit{iv)} high quality documents and Python libraries will be made and added for easier access of the existing and new resources.
%cases with the issue of lacking of labeled data, such as a new ZS-KGC task with unseen entities and problems from other domains e.g., zero-shot visual question answering.
%When extending to broader domains, especially to some cases which are more specialized or not well-documented, the accessibility of knowledge sources is also in our concerns.
%Targeting this, we can consider reusing the resources we have created and treating them as a priori for other cases.

\section{Conclusion}\label{discussion}

External knowledge plays a critical role in ZSL, while KGs have shown their great superiority for representing different kinds of external knowledge for augmenting ZSL.
To address the issue of semantic insufficiency in existing ZSL resources and the issue of lacking standard benchmarks to investigate and fairly compare KG-based ZSL methods, 
we created systematic resources for KG-based research on ZS-IMGC, ZS-RE and ZS-KGC, including six standard ZSL datasets and their corresponding KGs that can support different settings with ranging semantics.
For ZS-IMGC, we integrate not only typical external knowledge such as class hierarchy, attributes and text, but also common sense relational facts from ConceptNet and some logical expressions such as class disjointness.
For ZS-RE, we contributed KGs equipped with logic rules as the external knowledge.
% of relation types
For ZS-KGC, we build ontological schemas with semantics defined by RDFS and OWL, such as relation hierarchy, relation's domain and range, concepts, relation characteristics, and relation and concept textual meta data, for a NELL KG and a Wikidata KG that are to be completed.
Based on these resources, we conducted an extensive benchmarking study on different ZSL methods under different external knowledge settings, which illustrate the effectiveness and great potential usage of our proposed resources. 
We also discussed the strongness and weakness of different KG-based methods of different paradigms, and analyzed potential solutions for addressing some specific knowledge settings, for adaption to different datasets across tasks, and for better exploiting different kinds of external knowledge.
%In the future, we consider extending our resources to support broader studies.
%Besides, as the zero-shot problems we investigated here mainly focus on a set of pre-defined classes or relations, we also consider constructing ZSL resources for more real-world situations where new unseen classes arise incrementally or even new seen classes and new training data are collected incrementally (i.e., incremental ZSL \cite{wei2021incremental}).

% \todo{[Provide an extensive list on what interesting tasks that researchers can use our benchmarks to do]}

% Uncomment and use as the case may be
%\begin{theorem} 
%\end{theorem}

% Uncomment and use as the case may be
%\begin{lemma} 
%\end{lemma}


\printcredits

\balance
% \bibliographystyle{cas-model2-names}
\bibliographystyle{elsarticle-num}

\bibliography{cas-refs}

\bio{}

\endbio

\end{document}